\documentclass[lettersize,journal]{IEEEtran}
\usepackage{amsmath,amsfonts}
\usepackage{amssymb}   
\usepackage{algorithm}
\usepackage{algorithmic}
\usepackage{array}
\usepackage[caption=false,font=footnotesize,labelfont=rm,textfont=rm]{subfig}
\usepackage{textcomp}
\usepackage{stfloats}
\usepackage{url}
\usepackage{verbatim}
\usepackage{graphicx}
\usepackage{cite}
\usepackage[colorlinks,
            linkcolor=blue,
            anchorcolor=blue,
            citecolor=red]{hyperref}
\def\BibTeX{{\rm B\kern-.05em{\sc i\kern-.025em b}\kern-.08em
    T\kern-.1667em\lower.7ex\hbox{E}\kern-.125emX}}
\usepackage{balance}

\begin{document}
\title{A Tightly Coupled Bi-Level Coordination Framework for CAVs at Road Intersections}
\author{Donglin Li, Tingting Zhang~\IEEEmembership{Member,~IEEE}, Jiping Luo, Tianhao Liang, Bin Cao~\IEEEmembership{Member,~IEEE}, Xuanli Wu~\IEEEmembership{Member,~IEEE} and Qinyu Zhang~\IEEEmembership{Senior Member,~IEEE}
\thanks{D. Li, T. Zhang, J. Luo, T. Liang, B. Cao and Q. Zhang are with the School of Electronics and Information Engineering, Harbin Institute of Technology, Shenzhen. X. Wu is with the Communication Research Center, Harbin Institute of Technology, Harbin, P. R. China.

T. Zhang, B. Cao and Q. Zhang are also with Network Communication Research Center, Peng Cheng Laboratory, Shenzhen, P. R. China.

Email: zhangtt@hit.edu.cn}}


\maketitle

\begin{abstract}
Since the traffic administration at road intersections determines the capacity bottleneck of modern transportation systems, intelligent cooperative coordination for connected autonomous vehicles (CAVs) has shown to be an effective solution. In this paper, we try to formulate a Bi-Level CAVs intersection coordination framework, where coordinators from High and Low levels are tightly coupled. In the High-Level coordinator where vehicles from multiple roads are involved, we take various metrics including throughput, safety, fairness and comfort into consideration. Motivated by the time consuming space-time resource allocation framework, we try to give a low complexity solution by transforming the complicated original problem into a sequential linear programming one.
Based on the ``feasible tunnels'' (FT) generated from the high-Level coordinator, we then propose a rapid gradient-based trajectory optimization strategy in the low-level planner, to effectively avoid collisions beyond high-level considerations, such as the unexpected pedestrian or bicycles. Simulation results and laboratory experiments show that our proposed method outperforms existing strategies. Moreover, the most impressive advantage is that the proposed strategy can plan vehicle trajectory in milliseconds, which is promising in real-world deployments. A detailed description include the coordination framework and experiment demo could be found at the supplement materials, or online at https://youtu.be/MuhjhKfNIOg.

\end{abstract}


\section{Introduction}
\subsection{Background and Motivation}
With the rapid development of modern industry and urbanization, more and more vehicles are entering the transportation system, which adds great pressure to the current road infrastructure. In most developed areas, there exists rather limited space to build extra flyovers and highways, to accommodate the rapidly increased number of vehicles. Meanwhile, many transportation related issues such as the safety, traffic congestion, fuel consumption, air pollution have become social problems, and received tremendous attention from various departments so far \cite{Alsabaan2013,Dornbush2007,Shinde2017,Zhao2019,Wu2011}.

Since multiple roads merge at the road intersection, it has become one of the bottlenecks in modern transportation systems straightforwardly. According to reports in \cite{Choi2010CrashFI}, most traffic congestion and accidents occur at road intersections. Traditional intersection coordination strategies include traffic lights, roundabouts, stop signs, etc. However, 90\% of traffic accidents were related to the improper behaviors of human drivers, such as the dangerous or fatigue driving, negligence of traffic coordination signals or sudden appearance obstacles \cite{Bellis2008NationalMV}. Therefore, it is of vital importance to design dedicated intersection management systems to improve traffic throughput, as well as ensuring the driving safety \cite{Chen2016}.

With the rapid development of vehicular networking technology in recent years, vehicle-to-vehicle (V2V) and vehicle-to-infrastructure (V2I) communications \cite{Rawat2014,Vershinin2020} are no longer the bottleneck for multi-vehicle cooperation issues. Several studies \cite{Lin2017,li2019,Yao2023} have showcased the immense potential of CAVs at signal-free intersections. In comparison to studies on traditional single-vehicle ego planners \cite{motion2016}, this research explores more complex multi-vehicle planning problems. Coordinators at road intersections need to address two major issues. First, CAVs must be carefully coordinated to improve traffic throughput. Second, CAVs must perform ego motion planning based on environmental sensing to avoid collisions with accidental non-CAV obstacles, such as pedestrians, bicycles, and other non-connected vehicles.

\subsection{Related Work}

Currently, investigations on the cooperative coordination of CAVs at road intersections can be categorized into two trends. The first trend focuses on intersection modeling and vehicle scheduling strategies, which are usually referred to as ``centralized'' or ``high-level'' planner.
Another trend is the emergence of a bi-level structure built upon the centralized scheduling method. It is common for individual vehicle autonomous planners to get stuck in deadlocks \cite{Perronnet2019} due to system errors. To address this issue, some authors have proposed combining the results of a high-level planner with the autonomous planner for obstacle avoidance. This approach effectively ensures system robustness and traffic safety. The former emphasizes overall coordination efficiency and computational speed, while the latter emphasizes system robustness.

These high-level planners typically possess the ``global information'' of the entire intersection and can be implemented on road-side units (RSUs) or other infrastructure. One early investigation, as described in \cite{Lee2012}, prohibited any two CAVs from entering the intersection simultaneously to avoid collisions. Besides safety, this coordinator also considered other metrics such as traffic throughput and fuel efficiency. This work utilized a linear maneuver model with fixed acceleration, simplifying the system formulation significantly.

To further enhance throughput, Kamal \textit{et al.} introduced the concept of cross-collision points (CCPs) at road intersections \cite{Kamal2015}. Based on CCPs, Li \textit{et al.} proposed a cooperative critical turning point-based decision-making algorithm \cite{Li2022}, which modeled the coordinator as a partially observable Markov decision process (POMDP) problem to obtain the velocity of the critical turning points. However, the resulting velocity profiles were discontinuous and couldn't be realized using practical vehicle kinematic models. Additionally, this strategy faced high computational complexity.

Other typical high-level coordinators can be found in \cite{Hult2016,henk2017} and related references. These studies developed a strategy based on collision sets (CS), which is representative to formulating CS coordination as a mixed-integer programming (MIP) problem and solving it by introducing auxiliary variables\cite{Wang2018}. Since a single CS solution was not efficient, efforts were made to extend it to a multi-collision sets (MCS) model to improve traffic capacity with a certain increase in computational time \cite{Mo2018, Liu2019}. Moreover, some researchers also considered the ``stability'' of the intersection when vehicles continuously enter the intersection queue \cite{liu2020}, a topic rarely discussed before. What's more, Zhang\cite{zhangicas2021} provided a novel solution approach for the single CS strategy and incorporates local obstacle avoidance into the coordination of intersection control. However, in these investigations, vehicles adopted linear maneuver models, limiting the exploitation of the two-dimensional intersection's spatial resources. In \cite{Pan2023}, the authors incorporated friction losses and power train into the kinematic models to achieve more realistic solutions. The authors of \cite{Luo2022} proposed a method based on Deep Reinforcement Learning (DRL). This method fixes the lateral path of the vehicle and generates a speed profile. However, due to the fact that learning-based methods require a pre-determined model, this approach is not suitable for dynamic traffic environments (variable vehicle size and dynamics models). Additionally, in scenarios with a large number of vehicles, the efficiency of traffic flow can be compromised by vehicle grouping.

The second trend focuses more on system robustness, and therefore, researchers often propose a more robust double-level framework that extends the capabilities of the low-level planner for local obstacle avoidance \cite{zhangicas2021}. This approach leaves some safety redundancy and effectively prevents deadlocks. To develop a more efficient coordination strategy, integrated double-level coordination frameworks were proposed in \cite{zhang2021,Chen2021,Cong2022,Malikopoulos2021}. The high-level planner generates reference trajectories and corresponding feasible tunnels, while the low-level planner, similar to single-vehicle planners, avoids nearby non-CAV obstacles within the generated FT. A novel space-time resource searching (STRS) strategy was introduced in the high-level planner. However, this method suffers from the drawback of slow computation in the high-level planner, as it involves complex dynamic programming (DP) and quadratic programming (QP). Additionally, the efficiency of the low-level planner is comparatively low, requiring a substantial number of feasible tunnels for searching viable trajectories, due to insufficient coupling with the high-level planner.
Furthermore, the authors suggest that the deep reinforcement learning-based methods proposed in \cite{Luo2022} could also be extended to enhance the capabilities of the low-level planner \cite{{motion2016,fan2018}} in preventing local collisions caused by errors in localization and control systems.

In addition to the aforementioned study, one major problem faced in coordination is the presence of mixed traffic flow, where both Connected Autonomous Vehicles and Manual Vehicles (MVs) coexist. In this scenario, the difficulty of overall coordination is increased due to the uncontrolled nature of MVs. Consequently, some researchers have focused on addressing this issue. Kamal \textit{et al.} \cite{Kamal2020} proposed an adaptive traffic light control scheme, where the authors divided the intersection into several non-conflicting phases based on vehicle strategies. They allocated a non-conflicting green period to each vehicle according to its arrival status at the intersection. However, this method did not fully utilize the temporal and spatial resources within the intersection. Building upon the above method, Huang \textit{et al.} \cite{Huang2023} further refined the sapce-time resources within the intersection and grouped CAVs and MVs separately, introducing two specific lane restrictions. This method improves the rationality and efficiency, providing great inspiration for handling mixed traffic flow scenarios.

Some recent studies also suggest that dividing the global vehicle coordination problem into two-stages is a promising approach, namely decoupling the vehicle passage order from the specific trajectory optimization problem. For instance, in the work of Yao \cite{Yao2022,Yao2023}, a two-stage collaborative framework for Schedule and Trajectory Optimization is proposed. This framework first employs Mixed Integer Linear Programming (MILP) based on vehicle information to determine the optimal arrival time at intersections, and then solves the trajectory optimization problem based on the results from the scheduling phase. This method effectively reduces vehicle passage delays and improves fuel efficiency. In addition, Jiang \cite{Jiang2023} further consider the time-optimal vehicle passage order. The authors employ a heuristic Monte Carlo Tree Search (MCTS) method to accelerate the solving efficiency. It is worth noting that the authors innovatively explore the impact of all-direction turn lanes (ADTL) and specific-direction turn lanes (SDTL) on the scheduling algorithm.

However, the current research in the field urgently needs a fast and tightly coupled framework that can achieve obstacle avoidance in intersection scheduling, while ensuring high robustness and efficiency. In this case, tight coupled means that the low-level planner must heavily rely on the results of the high-level planner to ensure its effectiveness. The trajectories generated by the low-level planner should aim to avoid obstacles as closely as possible to the results of the high-level planner, reducing the probability of deviating from reference trajectories of high-level planner. Excessive deviation from the reference trajectory generated by the high-level planner can potentially lead to schedule breakdown. However, the ability of such tight coupled is not adequately demonstrated in \cite{zhang2021}. 
This framework should integrate high-level and low-level planners, taking into account the limitations of existing methods, such as the assumption of identical vehicle sizes, ideal communication, and control systems. It should also address the challenges posed by obstacles and control/positioning errors. To achieve high robustness and efficiency, a coordinated scheduling framework for signal-free intersections is needed.

\subsection{Main Contributions}
Aiming at the issues on current investigations, we try to propose a low-complexity and tightly coupled bi-level coordination framework for CAVs at non-signalized road intersections.
In the high-level planner, the traffic throughput, fairness and complexity are still main considerations. Unlike existing time exhausting investigations in \cite{zhang2021}, our method could significantly reduce the computing efforts. Furthermore, unlike the most existing {\it single level} coordinators, a tightly coupled low-level planner is proposed to handle non-CAV obstacles, which are not considered in the high-level planner. The main contributions can be summarized as follows.

\begin{itemize}
\item We propose an computational efficient solution to indicate the feasible tunnels for each CAVs, which has ability to extended to various kinematics constraints and scales of CAVs. Furthermore, considering the {\it fairness} and the {\it duration of scheduling}, we give a novel benchmark of passing order priority to determine traffic sequence of vehicles.

\item To ensure driving safety within the `` Feasible tunnel'', we propose an effective motion plan method based on the gradient and B-spline curve. It generates the initial control points of the B-spline curve, then optimizes the control points for lower smoothness cost and avoids potential collision under the feasible tunnel constraints. Comparing to existing solutions, the tightly coupled local motion planner is more suitable to the double-level coordination framework, which could make full advantage of the result of high-level planner and higher computational efficiency.

\item The simulation and laboratorial experiments show that our proposed approach can effectively improve the traffic throughput. One excellent advantage is that our high-level planner has only millisecond-level latency, which means that the strategy can be deployed in real-world settings.
\end{itemize}

The rest of this paper is organized as follows: In Section \ref{sec::systemModel}, we will explain the intersection model of our framework, vehicle kinematic model, the collision detection, construction of our framework and some state-of-art benchmark solutions for comparison. In section \ref{sec::highLevel}, the modified coordination strategy based on space-time resource blocks is introduced. In section \ref{sec::lowLevel}, a novel motion planner based on gradient is proposed, which is more suitable for the double-level framework. The simulation and laboratorial experiments results are demonstrated in Section \ref{sec::simutaion} and Section \ref{sec::LabExp}, respectively.

\section{Preliminary}
\label{sec::systemModel}

\subsection{Intersection model}
\begin{figure}[htb]
	\centering
	\includegraphics[width=0.7\columnwidth]{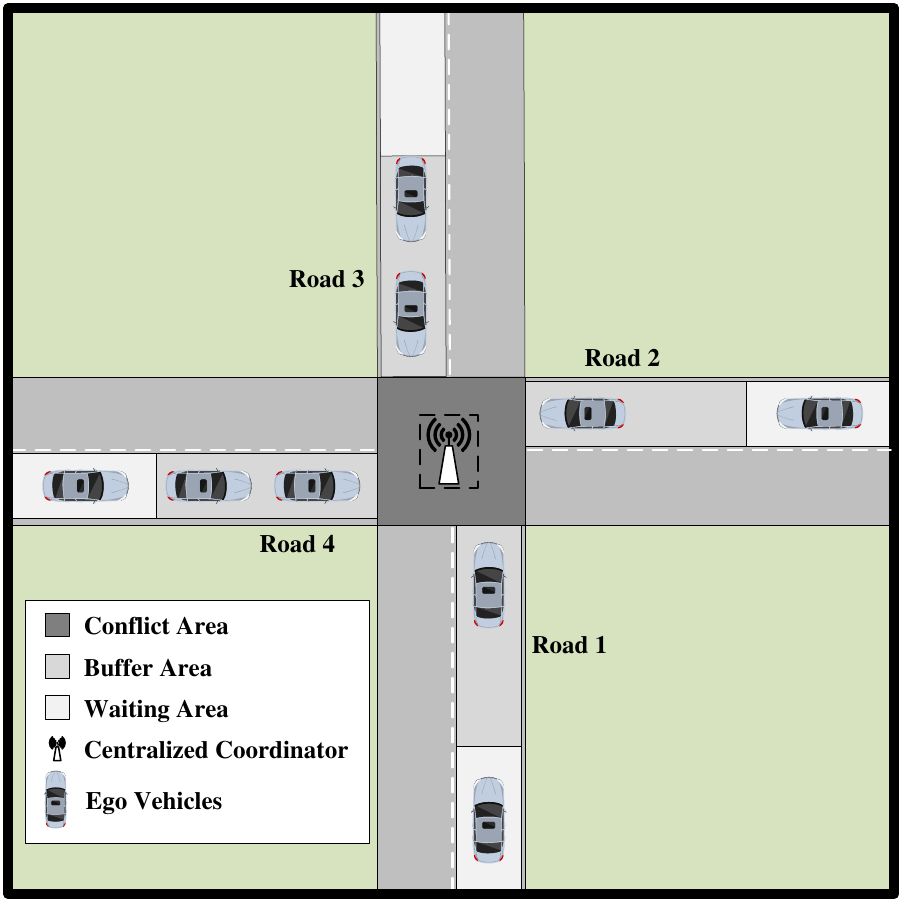}
	\caption{System model of the intersection.}\label{fig::intersection}
\end{figure}

We deploy a typical road intersection which consists of $R$ roads (for easy illustration, we set $R = 4$), as shown in Fig. \ref{fig::intersection}. The road intersection can be divided into three main sections, {\textit{i.e.}}, Waiting Area (WA), Buffer Area (BA) and Conflict Area (CA). The CA is the crucial area in our coordination framework, where potential collisions may occur. The width of lanes is $W_L$, the length of BA is $L_B$. 

We use $\mathcal{R} = \{1,2,...,R\}$ and $\mathcal{I}_{r} = \{1,2,...,I_{r}\}$ to represent the sets of roads and CAVs in road $r$, respectively.
At signal-free intersections, the decision making center may face high dynamic scene, {\textit{e.g.}}, the unexpected obstacles, unsatisfactory communication and control systems. It is necessary to make following assumptions:
\begin{enumerate}[]
  \item Each CAVs will send its state information (position, velocity, etc.) and intention maneuvers ({\textit{i.e.}}, go straight, turn left and turn right) to the CC via vehicle-to-infrastructure links, when entering the WA. After executing the coordination framework, the CC will broadcast the coordination information ({\textit{i.e.}} the moment of entering intersection and feasible tunnels).
  \item After received the coordination information, the CAVs should adjust their velocity in BA to enter the intersection at the timing that broadcasted by CC. Then CAVs travel along with the reference trajectory which is predefined, if there is no unexpected obstacles.
  \item Without loss of generality, the high-level planner will allocate the space-time resource blocks to every CAVs. Considering the unexpected obstacles, there should be some reasonable redundancy involved by the feasible tunnel to make sure the low-level planner has enough space to avoid collision.
\end{enumerate}

\begin{figure}[htb]
	\centering
	\includegraphics[width=0.83\columnwidth]{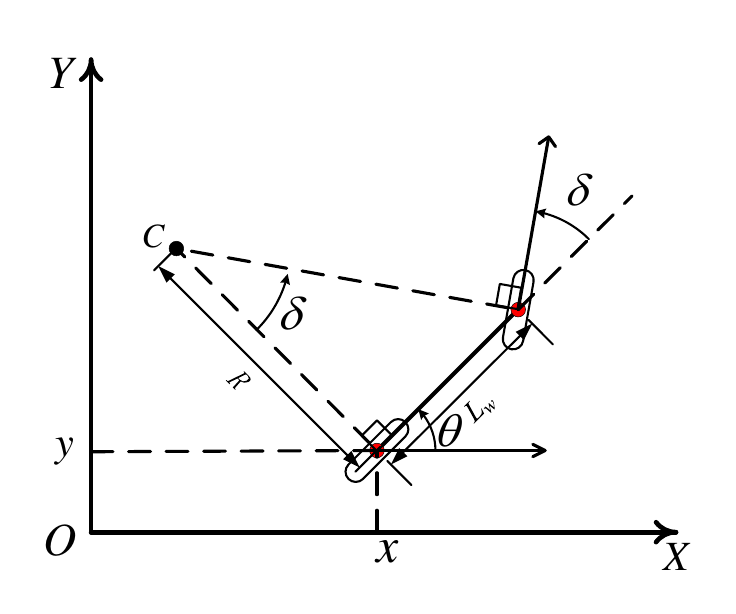}
    \vspace{-1em}
	\caption{The kinematic model of a front steering vehicle.}\label{fig::kinematicmodel}
\end{figure}

\subsection{Vehicle Kinematic Model}
In this paper, to reduce the computational complexity, we adopt the popular kinematic bicycle model \cite{kong2015} to describe the movement of vehicles, as shown in Fig. \ref{fig::kinematicmodel}. The kinematics state of the $i$-th vehicle in road $r$ can be described as $[x_{i,r}, y_{i,r}, \theta_{i,r}, v_{i,r}]^\text{T}$, where $(x_{i,r}, y_{i,r})$, $v_{i,r}$ and $\theta_{i,r}$ are the position, velocity and heading in 2D Cartesian coordinate system, respectively. During the actual driving behavior, the control input are the longitudinal acceleration and the steering angle, donated by $a_{i,r}$ and $\theta_{i,r}$, respectively. The vehicle kinematic state update equation is shown as following:

\begin{equation}
	 \begin{matrix}
		\left[ \begin{matrix}
			{\dot{x}_{i,r}}  \\
			{\dot{y}_{i,r}}  \\
			{\dot{\theta}_{i,r}}  \\
			{\dot{v}_{i,r}}  \\
		\end{matrix} \right]= \left[ \begin{matrix}
			v_{i,r}\cos (\theta_{i,r})  \\
			v_{i,r}\sin (\theta_{i,r})  \\
			v_{i,r}\tan{\delta_{i,r}}/L \\
			{a_{i,r}} \\
		\end{matrix} \right]
	\end{matrix} \ ,r \in {\cal R},i \in \mathcal{I}_{r} \label{equ::bicycleModel}
\end{equation}
where $(\dot{\bullet}):=\frac{\partial}{\partial t}(\bullet)$, $L_w$ is the distance between the front and rear wheel axles. What's more, taking into account the kinematic constraints of the vehicle, there must have:
\begin{equation}
\begin{aligned}
a^{\min }_{i,r} & \leq a_{i,r} \leq a^{\max }_{i,r} \\
0 & \leq v_{i,r} \leq v^{\max }_{i,r} \\
-\delta^{\max }_{i,r} & \leq \delta_{i,r} \leq \delta^{\max}_{i,r}
\end{aligned}
\end{equation}

\subsection{Space Time Resource Blocks}
The space-time resources was first used to describe the resources in the intersection by \cite{zhang2021}. The resources in CA are described by three-dimensional coordinates in the $X$, $Y$ and $T$ domains. For the convenience of computation and reasonability, the space-time resources are divided into multiple blocks with the resolutions $dx$ in $X$ domain, $dy$ in $Y$ domain, $dt$ in $T$ domain. Then the resource in $(x, y, z)$ could be described as:
\begin{equation}
\label{equ::resources}
    \mathbb{B}_{j_{x}, j_{y}, j_{t}}=\left\{
    \begin{array}{l}
    (x, y, t) \bigg| \begin{array}{l}
    \left(j_{x}-1\right) \cdot {d} x \leq x<j_{x} \cdot {d} x \\
    \left(j_{y}-1\right) \cdot {d} y \leq y<j_{y} \cdot {d} y \\
    \left(j_{t}-1\right) \cdot {d} t \leq t<j_{t} \cdot {d} t
    \end{array}
    \end{array}
    \right\}
\end{equation}
where $j_x, j_y$ and $j_t$ are the indexs in $X, Y$ and $T$ domains, respectively. One space-time can not be occupied by more than one vehicle simultaneously. Thus, when attempt to allocate the space-time resource in $(j_x, j_y, j_t)$, the Centralized Coordinator must check the occupancy state of the space-time resource blocks.

The allocated area for the vehicle is represented as a collection of resource blocks, which includes not only the resource blocks occupied by the vehicle, with proper redundancy.
We define the resource occupied by the $i$-th vehicle of road $r$ as
\begin{equation}\label{equ::occupancy}
  \mathbb{A}_{i, r}=\bigcup \mathbb{B}_{j_{x}, j_{y}, j_{t}} \bigg |_{(j_x, j_y, j_t) \in \mathcal{A}_{i, r} }
\end{equation}
where $\mathcal{A}_{i, r}$ represents the index set of resource blocks occupied by $i$-th vehicle of $r$-th road. To keep no collisions among vehicles, we need to ensure that one resource block can only be allocated to one vehicle. That means the intersection of allocated resources between the two vehicles is empty, {\textit{i.e.}},
\begin{equation}\label{equ::collisionFree}
  \mathbb{A}_{i1, r1} \bigcap \mathbb{A}_{i2, r2} = \emptyset
\end{equation}
with $\forall r1,r2 \in \mathcal{R}$, $\forall i1 \in \mathcal{I}_{r1}$ and $\forall i2 \in \mathcal{I}_{r2}$.


%
%

\subsection{Collision Detection}

\begin{figure}[htb]
	\centering
	\includegraphics[width=0.83\columnwidth]{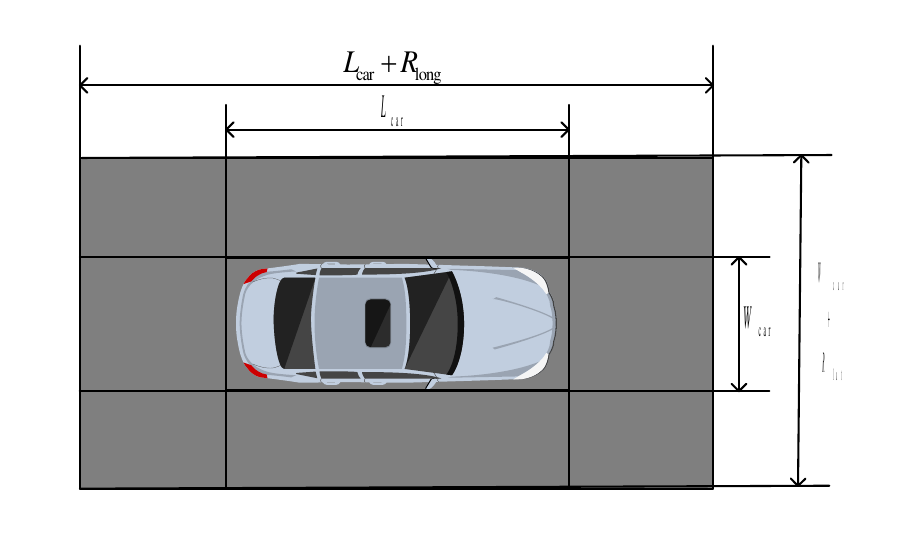}
	\caption{The occupied region of a vehicle with safe redundancy}\label{fig::redundancy}
\end{figure}

In practice, there are noises in the localization \cite{Liang2022} and control systems,
\begin{equation}\label{equ::localError}
    \left\{ \begin{matrix}
     x_\text{m} \sim \mathcal{N}(x,\sigma_x^2)\\
    y_\text{m} \sim \mathcal{N}(y,\sigma_y^2)\\
    \theta_\text{m} \sim \mathcal{N}(\theta,\sigma_{\theta}^2)\\
    \end{matrix}\right.
\end{equation}

\begin{equation}\label{equ::ctrlError}
    \left\{ \begin{matrix}
    a_\text{f} \sim \mathcal{N}(a,\sigma_a^2)\\
    \delta_\text{f} \sim \mathcal{N}(\delta,\sigma_\delta^2)\\
    \end{matrix}\right.
\end{equation}
where $(x,y,\theta)$ and $(x_\text{m}, y_\text{m},\theta_\text{m})$ are the actual values and measured value of localization systems, respectively. $(a, \delta)$ and $(a_\text{f}, \delta_\text{f})$ are the expected values and actual value of control systems, respectively.
The CAVs always deviate from the planned trajectory. These imperfections need to be treated carefully. For the safety of vehicles, we set a redundancy around the vehicle, as shown in Fig. \ref{fig::redundancy}. The $L_{\text{car}}$ and $W_{\text{car}}$ represent the length and width of vehicle, respectively. The $R_{\text{long}}$ and $R_{\text{lat}}$ are the longitudinal and lateral safe redundancy. The parameters of redundancy are decided by the density of obstacles and performance of control and localization system.

During the coordination process and local motion planning, the redundancy region is also considered when collision detection.

\begin{figure}[htb]
	\centering
	\includegraphics[width=0.6\columnwidth]{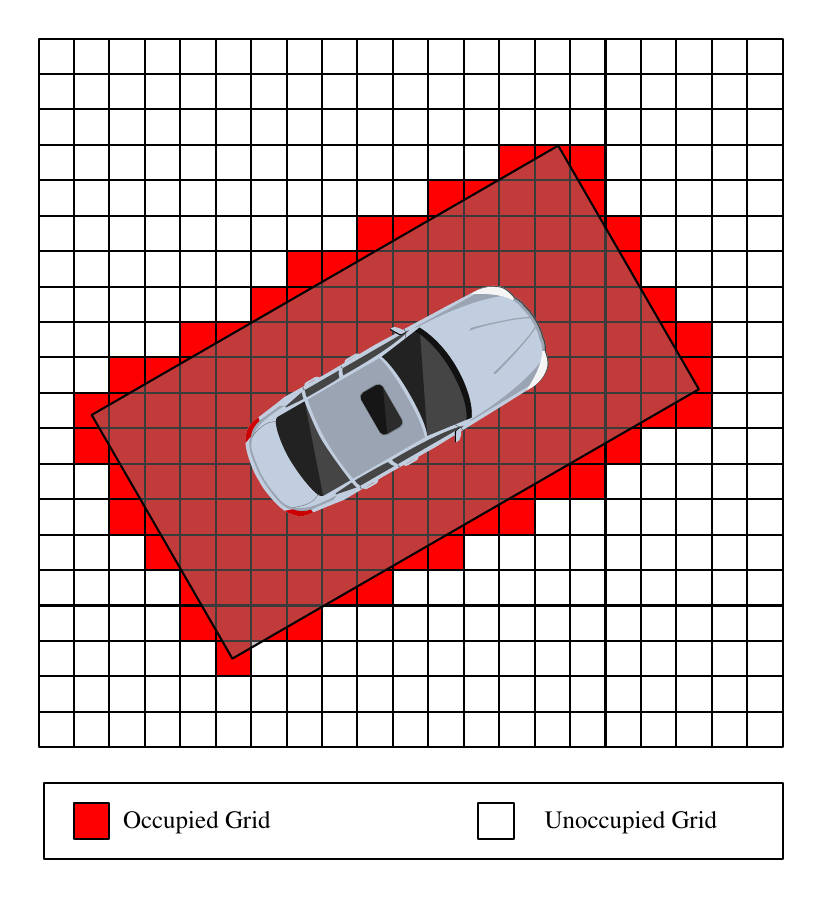}
	\caption{The occupancy grid map of a vehicle with safe redundancy}\label{fig::gridmap}
\end{figure}

The space-time resource blocks occupied by other CAVs in time $t$ is could be considered as a occupancy grid map \cite{li2015}, as shown in Fig. \ref{fig::gridmap}. For a grid that has been occupied, it could not be occupied repeatedly, otherwise a collision will occur. In low-level planner, we assumed that the CAVs could build a Object Oriented Bounding Boxes (OOBB) topological map with the equipped sensors, then transform the OOBB map into the occupancy grid map.

In theory, redundancy is a trade-off between coordination efficiency and system robustness. Consequently, we attempt to propose a rough estimation method for redundancy size, although it should be noted that it may not be precise, through theoretical analysis. Generally, we can choose the maximum value of three times the standard deviation of the lateral and longitudinal offset errors caused by vehicle localization and control system errors as a factor to consider. In addition, unexpected obstacles within the intersection should also be taken into account. We can calculate the diameter of the circumscribed circle outside the OOBB of each obstacles as another statistical factor, which can be observed by CC. Although the obstacles may not be located on the boundary of the feasible path, which may cause the vehicles to deviate from their intended trajectories to go around the obstacles, the system is able to handle such deviations to a certain extent due to the robustness of the low-level planner. Finally, we define a rough formula for calculating the redundancy size of feasible tunnel:

\begin{equation}
    \left\{ \begin{matrix}
        R_{\textrm{lat}}=\max(3\sigma_{\textrm{lat}}, d_{\textrm{OOBB}})\\
        R_{\textrm{long}}=3\sigma_{\textrm{long}} \\
       \end{matrix}\right.
\end{equation}
where $\sigma_{\textrm{lat}}$ and $\sigma_{\textrm{long}}$ are the variances of the lateral and longitudinal tracking errors due to the subpar localization and control system, respectively, $d_{\textrm{OOBB}}$ is the diameter of the circumscribed circle outside the bounding box of the obstacle.

In both High-Level and low-level planners, we calculate the grids that occupied by CAVs and obstacles, respectively. For collision free, there can not be any overlaps of allocated grids.

\begin{figure*}[htb]
	\centering
	\includegraphics[width=1.4\columnwidth]{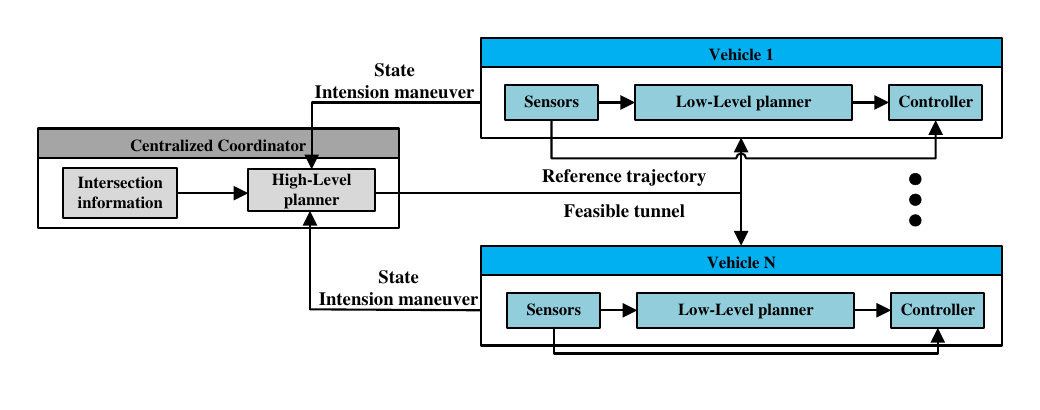}
    \vspace{-1em}
	\caption{The structure of coordination framework}\label{fig::structure}
\end{figure*}
\subsection{Structure of the Coordination Framework}
The double-level coordination framework can be divided into two parts, as depicted in Fig. \ref{fig::structure}. The functions of two parts are as following.

\begin{enumerate}[]
  \item high-level planner (coordination strategy): After the CAVs entering the WA, the CC attempts to receive the their information about positions, velocity, intention maneuvers and the like by V2I (Vehicle to infrastructure) links. According to road information of the intersection, the CC generates the reference trajectories for each CAV. After calculated the traffic priority of the first CAVs in each road, the CC executes coordination strategy to get the reference trajectories and feasible tunnels of the waiting CAVs.

  \item Low-level planner (local motion planner): Based on the reference trajectory generated by the high-level planner, the low-level planner attempts to optimize the trajectory to avoid collisions within the feasible tunnel. The practice trajectory also needs to be sufficiently smooth and feasible in terms of kinematics model. What's more, an unexceptionable controller also plays an important role in the frame, and lower control errors can reduce the probability of collisions. It is worth noting that the low-level planner is executed by CAVs based on local perception information and, therefore, does not occupy resources of the central coordinator.
\end{enumerate}

\subsection{Benchmark Solutions} \label{sec:Benchmark}
In this subsection, we briefly introduce some state-of-art benchmark solutions of coordination strategy in intersection for comparison.

\subsubsection{Collision Set}
The area where exists potential collision is defined as a collision set in Collision Set strategy \cite{zhangicas2021}, that means one collision set can not be occupied by multiple CAVs simultaneously, the CAVs are only allowed to pass through the intersection one after the other. Although the CS strategy shows preponderance in terms of long-term stability and relative low-complexity implementation, if cannot tap the full potential of the intersection. The high-level planner $\uppercase\expandafter{\romannumeral1}$ in \cite{zhang2021}, the authors imporved the CS strategy for better performance. Thus, we regard the CS strategy proposed by \cite{zhang2021} as our benchmark for comparison.

\subsubsection{Space-Time-Block Searching}
\cite{zhang2021} proposed a novel and aggressive space-time resource searching strategy. For maximum pass throughput, multiple CAVs can share the intersection simultaneously. The space-time resource is divided into many blocks which can be regarded as three-dimensional coordinates with $X$, $Y$, and $T$ domains. Based on the reference path, CC searches the velocity profiles of CAVs in ST domains to guarantee comfort and safety. Due the transition of resources between XYT domains and ST domains, this strategy could get the optimal trajectory at the expense of high computation complexity. In comparison to CS strategy, STRS still shows a obvious throughput advantages. In addition, STRS strategy can not operate in real-time, and the computational complexity grows exponentially.


\section{high-level planner}
\label{sec::highLevel}

\begin{figure}[htb]
	\centering
	\includegraphics[width=0.83\columnwidth]{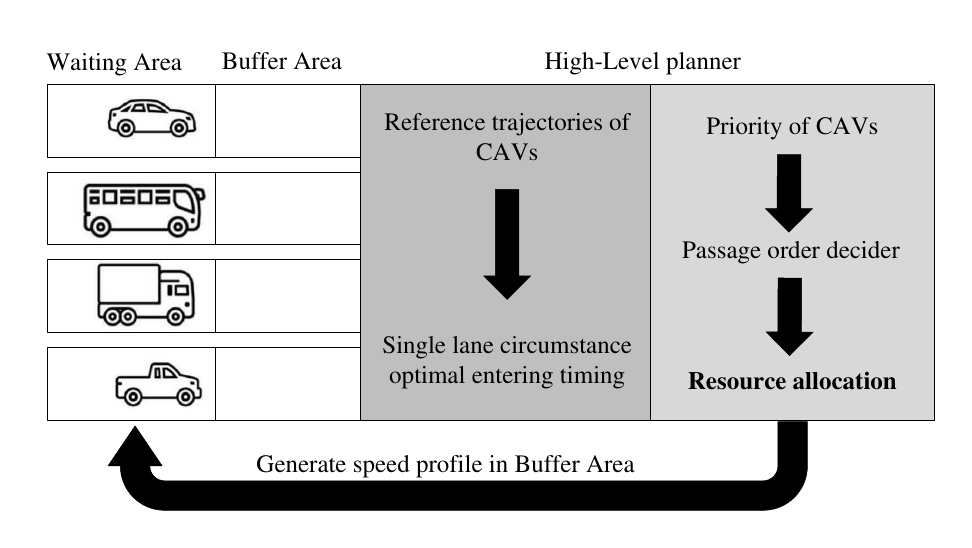}
	\caption{The structure of high-level planner}\label{fig::highLevel}
\end{figure}

The structure of the entire high-level planner is shown in Fig. \ref{fig::highLevel}.
Firstly, we generate reference trajectories of different CAVs according to their kinematic constraints by Reference Trajectory Generator.
We can obtain the reasonable timing of CAVs to enter the intersection by the core module of our proposed, then the CC determines the current round passage CAV by comparing the priority of CAVs, and allocates its feasible tunnel to complete the current round of coordination.
As the result of the high-level planner is the timing of the CAV to enter the Conflict Area, the trajectory between the Waiting Area and the Conflict Area needs to be calculated finally. We'll introduce each of the modules mentioned above.

\subsection{Reference Trajectory Generator} \label{sec::RTG}
The simplified line $\&$ circle path \cite{Horst2006} is one popular option. However, it is too ideal to be realized in the kinematic bicycle model due to the discontinuous second order derivative.
In the three-dimensional Cartesian coordinates system XYT, the trajectory can be decoupled as Path and Speed Profile \cite{Werling2010}. Usually, the trajectory can be expressed as

\begin{equation}
\left\{\begin{array}{l}
\text { Path: }\left\{\begin{array}{l}
x=f\left(s\right) \\
y=g\left(s\right)
\end{array}\right. \\
\text { Speed Profile: } u=s(t)
\end{array}\right.
\end{equation}
where $u$ is the arc length along with the path, $s$ is decided by velocity and acceleration of vehicle.

In order to make the path smooth and subject to the kinematics constraints, the problem can be built as a constrained optimization problem, one key constraint is the continuity of trajectories \cite{zhangicas2021}.
\begin{equation}
\label{equ::constranint1}
\begin{array}{l}
f(0) = x_0,\; g(0) = y_0 \\
f(s_{t}) = x_{s_t},\; g(s_{t}) = y_{s_{t}}

\end{array}
\end{equation}
where $(x_0,y_0)$ and $(x_{s_t},y_{s_t})$ are the start and end position of standard line $\&$ circle path, respectively. $s_t$ is the total length of standard path.

In order to make the reference trajectory more consistent with traffic rules and human driving habits, the distance limitation between the actual path and the standard line $\&$ circle path is also a necessary constraint.
\begin{equation}
\label{equ::constranint2}
\begin{array}{l}
\left | f(s_j) - x_j \right | \leq d_{\text{b}} \\
\left | g(s_j) - y_j \right | \leq d_{\text{b}}
\end{array}, \;
\forall j \in \left \{ 1,2,\dots,N_s \right\}
\end{equation}
where $(x_j, y_j)$ are the equally spaced sample points on standard line $\&$ circle path. $N_\text{s}$ is the number of sample points, which is decided by $\Delta s$. $d_{\text{b}}$ is the distance limitation between reference and standard line $\&$ circle paths.

Therefore, the optimization problem can be expressed as
\begin{align}
    \mathcal{P}_1: \;
	\underset{}{\mathop {\min. }}  \quad & \sum_{i = 2}^{4}\omega_{i}^{\text{ref}}\int (f^{(i)}(s))^2 +(g^{(i)}(s))^2\text{d}s
    \\
	\text{s.t.}
	\quad&(\ref{equ::constranint1}),\;(\ref{equ::constranint2})\nonumber
\end{align}
where $\bullet^{(i)}(s)$ means the $i$-th order derivative of $\bullet(s)$.

Generally, the quintic polynomial is a good way to express the trajectory, then $\mathcal{P}_1$ can be formulated as a typical quadratic programming problem \cite{Mellinger2011}, and solved by off-the-shelf solvers rapidly.

Since the length of entire standard line $\and$ circle path $s_\text{t}$ will be changed after planning, we propose a method to generate speed profiles rapidly for the higher fuel efficiency \cite{Hult2016} and comfort. Ideally, we want the vehicle to cross the intersection with a speed close to the reference speed $v_{\text{ref}}$ without any accelerations. 
After solving the reference path, we can obtain the length of the reference path $s_n$, thus the longitudinal displacement can be expressed as
\begin{equation}
    s(t)=\frac{s_t}{s_n}v_{\textrm{ref}}t
\end{equation}
which is the integral of velocity, also means the longitudinal travel distance along the reference path.

\subsection{Problem Formulation}
The high-level planner of STRS strategy can not solute in real-time, because of the high computation complexity. In order to apply the double-level framework based on space-time resource block to multi-roads intersection, we improved the model to overcome the high computation complexity and propose a new high-level planner to achieve feasible tunnel allocation for higher efficiency in traffic utilization.

First, the CC should calculate the reference trajectory for each vehicle. Specially, the potential collision exists in the Conflict Area, so we only consider the reference trajectory between the vehicle and redundancy entering the Conflict Area and leaving the Conflict Area.
Then, the reference trajectory of $i$-th vehicle in road $r$ can be represented as following,
\begin{equation}
\begin{array}{l}
\text { Trajectory: }  \left\{\begin{array}{l}
x_{i,r}(t)=f_{r_\text{f}, r_\text{t}, m}(t - t_{i,r}^\text{e}) \\
y_{i,r}(t)=g_{r_\text{f}, r_\text{t}, m}(t - t_{i,r}^\text{e}) \\
\theta _{i,r}(t) = \arctan{\frac{f^{'}_{r_\text{f}, r_\text{t}, m}(t - t_{i,r}^\text{e})}{g^{'}_{r_\text{f}, r_\text{t}, m}(t - t_{i,r}^\text{e})}}
\end{array}\right. ,
\\
 0 \leq t - t_{\textrm{e}} \leq t_{r_\text{f}, r_\text{t}, m}
\end{array}
\end{equation}
where $r_\text{f}$ and $r_\text{t}$ are the road which the vehicle come from and intend to reach, respectively. Specifically, $m$ is the maneuver (turn left, go straight, turn right) of the vehicle, that means road $r_\text{t}$ must be one of the feasible roads, which vehicle came from $r_\text{f}$ road and maneuvered as $m$ could reach. $t_{i,r}^e$ and $t_{r_\text{f}, r_\text{t}, m}$ are the moments of entering the Conflict Area and the duration of the specified reference trajectory, respectively. We can calculate the index set $\mathcal{C}_{r_\text{f},r_\text{t},m}^t$ of resource blocks occupied by vehicle at time $t$ state directly.

Generally, the index set of occupied resource block of each trajectory starting at time $t^\text{e}$ are represented as
\begin{equation}\label{equ::reftrajocc}
  \mathcal{D}^{t^\text{e}}_{r_\text{f},r_\text{t},m} =
  \left \{
  (j_x, j_y, j_t+\left \lfloor \frac{t^\text{e}}{\text{d}t} \right \rfloor)
  \bigg |
  \bigcup_{0<t<t_{r_\text{f}, r_\text{t}, m}}
  \mathcal{C}_{r_\text{f},r_\text{t},m}^t
  \right\}
\end{equation}
where $\left \lfloor \bullet  \right \rfloor$ means the round down of $\bullet$. $\mathcal{C}_{r_\text{f},r_\text{t},m}^t$ means the index set of resource blocks which are occupied vehicle with $(r_\text{f}, r_\text{t}, m)$ state at time $t$. Because the starting time of reference trajectory provided by Reference Trajectory Generator is $0$, the time index of reference trajectory index set must be added with a time offset $\lfloor \frac{t_\text{e}}{dt} \rfloor$.

\subsection{Single Lane Circumstance}
When there's only one road need to be coordinated, CAVs could only enter the intersection one by one. Thus, we should indicate the timing of entering the intersection for each CAVs.

At $k$-th round, the index set of space-time resource blocks occupied in previous $1 \sim k- 1$ rounds can be expressed as $\mathcal{A}_{\text{pre}}^k$. Particularly, $\mathcal{A}_{\text{pre}}^1 = \emptyset$ in the first round. For potential collisions avoidance, we need to guarantee that the same resource block cannot be occupied repeatedly, {\textit{i.e.}},
\begin{equation}\label{equ::constraint1}
  \mathcal{D}^t_{r_\text{f},r_\text{t},m} \bigcap \mathcal{A}_{\text{pre}}^k = \emptyset
\end{equation}

At $k$-th round, the resource that allocated to the $i$-th CAV in $r$-th road can be represented by $\mathcal{A}_{i, r}^k = \mathcal{D}^t_{r_\text{f},r_\text{t},m}$. After $k$-th round, $\mathcal{A}_{\text{pre}}$ should be updated.
\begin{equation}\label{equ::resourceupdate}
  \mathcal{A}_{\text{pre}}^{k+1} =  \mathcal{A}_{i, r}^k \bigcup \mathcal{A}_{\text{pre}}^k
\end{equation}

Thus, the resource allocation problem can be described as a typical linear programming problem:
\begin{align}
    \mathcal{P}_2: \;
	\underset{}{\mathop {\min.}}  \quad &\max. \left \{j_t | j_t \in \mathcal{A}_{\text{final}} \right \}
    \\
	\text{s.t.}
	\quad&(\ref{equ::collisionFree})\nonumber
\end{align}
where $\mathcal{A}_{\text{final}}$ represents the index set of resource blocks occupied during the whole scheduling process.

The aforementioned method can determine the optimal timing for a CAV to enter the intersection in a single-lane scenario. However, in a multi-lane intersection, there are multiple lanes to consider, which means we must select a CAV to enter the intersection first and then consider the subsequent CAVs in the next round of coordination. This implies that the problem can be solved using sliding window method, also means receding horizon control.

\subsection{Priority of CAVs}
Due to the potential presence of multiple vehicles waiting to pass at an intersection, it is necessary to determine the sequence of vehicle movements, which signifies the need to assign a priority, known as right of way, to each vehicle. That implies the optimization objective of the problem will change. We need to consider multiple optimization objectives, rather than solely focusing on traffic efficiency.
First and foremost, based on the resource blocks allocated to CAV, the passing priority term of traffic efficiency $P_d$ could be deﬁned as

\begin{equation}\label{equ::pd}
P_{\text{d}}=w_{\text{d}} \cdot {\max \left\{j_{t} \mid j_{t} \in\left(\mathcal{A}_{i, r}^k \cup \mathcal{A}_{\text{pre}}^k \right)\right\} \cdot {d}t}
\end{equation}

The general intersection coordination strategy all aim to maximize throughput, that can lead to unfair situations where the first CAV at some roads cannot be scheduled for a long time.
\cite{zhang2021} also ensures the fairness of the passage by introducing resource occupancy rate term into the performance criteria of the coordination. Nevertheless, vehicles going straight tend to increase resource occupancy rate more than turning left in most cases. In practice, the reasonableness of this item is debatable.
An effective approach is to add waiting time term to the performance criteria of the coordination for fairness of passage.

\begin{equation}\label{equ:pw}
P_{\text{w}}=\omega_{\text{w}} \cdot \sum_{n = 1}^{i} \frac{t_{\text{waiting}}^n}{n}
\end{equation}
where $t_{\text{waiting}}^n$ indicates the waiting time at $n$-th in road queue.

%

For queue stability \cite{Neely2010Stochastic}, according to Lyapunov theorem, we should ensure the following equation holds,
\begin{equation}\label{equ:queueStable}
  \lim_{T\rightarrow \infty } \sup \frac{1}{T}\sum_{t=1}^{T}E \left [ l_\text{r}(t) \right ] < \infty, r\in R
\end{equation}
where $l_\text{r}(t)$ is the length of $r$-th road at $t$.

Thus the performance criteria about queue stability can be defined as:
\begin{equation}\label{equ:psta}
P_{\text{sta}}=  \frac{\omega_{{\text{sta}}} \cdot l_\text{r}(t)a_\text{r}^{\text{av}}}{\sum_{i\in \mathcal{R}}^{}l_{i}(t)}  
\end{equation}
where $a_\text{r}^{\text{av}}$ represents the arrive rate of $r$-th road.

In every round, we could select the optimal vehicle $v$ in all road to pass base on priority of vehicle function, {\textit{i.e.}},
\begin{equation}\label{equ:p}
    v = \text{arg} \min \left \{ P_{\text{d}} + P_{\text{w}} - P_{\text{sta}} \right \}  
\end{equation}
where, we neglected the inclusion of the $P_{occ}$ factor proposed in \cite{zhang2021} due to its high coupling with $P_d$ factor. Specially, $v$ must be the first vehicle in all road queue in this round.

The algorithm flow of the entire high-level planner is shown in Algorithm \ref{alg:highLevel}.

\begin{algorithm}[htb]
\caption{high-level planner}
\label{alg:highLevel}
\begin{algorithmic}[1]
\STATE {Initialization. Set initial CAVs $\forall r\in\mathcal{R}, \forall i\in\mathcal{I}_r$, set $k = 1$ and
$\mathcal{A}_{\text{pre}}^1 = \emptyset $}.
\STATE{Get the information of intersection and vehicle, generation standard reference trajectories}.

\REPEAT
    \FOR {each $r \in R$}
    \STATE{Calculate the priority of first vehicle in road according \ref{equ:p}}.
    \ENDFOR
    \STATE{Select the optimal vehicle $v$}.
    \STATE{$j_{t} = 0$}.
    \REPEAT
        \STATE{$j_t= j_t + 1$}.
    \UNTIL{$\mathcal{D}^{j_t}_{r_\text{f},r_\text{t},m} \bigcap \mathcal{A}_{\text{pre}}^k = \emptyset$}
    \STATE{Allocate resource for vehicle $v$, {\textit{i.e.}}, $\mathcal{A}^k_{i,r} = \mathcal{D}^{j_t}_{r_\text{f},r_\text{t},m}$}
    \STATE{Update allocated resource, $\mathcal{A}^{k+1}_{\text{pre}} = \mathcal{A}_{i, r}^k \bigcup \mathcal{A}_{\text{pre}}^k$ }
    \STATE {$k = k + 1$}.
\UNTIL{The co-scheduling process is complete}.
\end{algorithmic}
\end{algorithm}

\section{low-level planner}
\label{sec::lowLevel}
For CAVs, there exist various random obstacles, such as pedestrians, bicycles, etc. They usually cannot be accurately tracked or predicted by road infrastructures, so such obstacles cannot be considered in the high-level planner. In addition, due to imperfect control and localization operations, CAVs may move out of their feasible tunnel and collide with other CAVs.
The CAVs need to characterize obstacles and the environment to generate the actual trajectory based on the reference trajectory and feasible tunnel provided by high-level planner.

The reference trajectory provided by high-level planner is optimal in terms of smoothness, since Reference Trajectory Generator was modeled as a quadratic programming problem. The optimality and feasibility of the reference trajectory will be destroyed due to the consideration of obstacles. The existing approach is to discard the original reference trajectory and recalculate the path and speed profile in the convex feasible tunnel. Obviously, recalculating the trajectory with DP + QP strategy is extremely time challenging.

Theoretically, the coordination efficiency of the high-level planner is closely related to the redundancy size, and the DP+QP method generally uses a circular collision detection model, which leads to large and inefficient redundancy, so a fast and efficient low-level planner is necessary to be proposed.

\subsection{Trajectory connection and stitching} \label{sec::TBA}
Due to the vehicle's kinematic constraints, usually, the vehicle cannot instantaneously accelerate to a specific velocity. Therefore, we must ensure the continuity between trajectories, which means that the generated trajectory by the planner must be continuous with the current trajectory at a specific time point in order to guarantee seamless trajectory connection and switching. After the vehicle enters the buffer area of the intersection, the vehicle's trajectory needs to switch from its own planner to a dedicated low-level planner for the intersection. The low-level planner requires a reference path, which means that we need to generate a continuous and smooth trajectory for trajectory planning. As we aim to ensure third-order continuity of the trajectory, {\textit{i.e.}}, continuity of position, velocity, acceleration, and jerk, a fifth-degree polynomial is undoubtedly a good representation. Since the general planner operates iteratively and the previous frame trajectory ensures collision-free, sampling the points on the previous frame trajectory as the starting point for the next frame trajectory is effective. Based on the desired trajectory length, we can sample the endpoint on the reference trajectory. Similar to the Reference Trajectory Generator, we can also solve a quadratic programming problem to obtain one or multiple segments of quintic polynomials as the initial values for the low-level planner. Due to space limitations, a detailed introduction is not provided here.

\subsection{Problem Formulation}
Inspired by \cite{Zhou2020}, we propose a gradient optimization-based trajectory generation scheme. In addition to the reference trajectory, we further consider four more factors: smoothness, kinematic feasibility, energy efficiency and collision avoidance, which can effectively avoid the time consuming trajectory re-calculation.
The trajectory is represented by a $p_\text{d}$-th order uniform B-Spline curve, and can be optimized by the control points of the B-Spline curve $\mathbf{Q}$ by gradient of the cost function. The original control points of B-Spline curve can be calculated by reference trajectories.

In general, the maneuver smoothness can be characterized via the acceleration, jerk and snap \cite{Mellinger2011}, specifically, the snap can be ignored when $p_\text{d} \leq 3$. The kinematics feasibility cost is represented by the limitation of lateral acceleration.
In terms of fuel efficiency, the optimal velocity profile is a constant $v_{\text{ref}}$ \cite{Hult2016}. Intuitively, we try to minimize the longitudinal acceleration, to approach the expected constant velocity. Thus, the smoothing, kinematics feasibility and energy efficiency terms of cost function can be expressed succinctly as following:
\begin{equation}
\label{equ:cost1}
C_{\text{s}}+C_{\text{k}}+C_{\text{e}} = \sum_{i = 2}^{3}\omega_{i}\int \left \| \mathbf{h}^{(i)}(t) \right \|^2_2 dt
\end{equation}
where $\mathbf{h}(t) = [f(t)\; g(t)]$.

The B-Spline curve is convex, which indicates that a span B-Spline curve can be controlled by $p_\text{d} +1$ successive control points. Another property is that the $k$-th order derivative of a B-Spline curve is also a B-Spline curve with $p_{\text{d},k} = p_\text{d} - k$. In our study, $p_\text{d} = 3$, the 3-th order derivative of a single span B-Spline curve is a constant, thus, the snap term can be ignored in the cost function \cite{Usenko2017}.

For uniform B-Spline curve with $N_\text{c}$ control points, the number of knots is $p_\text{d} + N_\text{c} + 1$, the knots vector can be represented by $\mathbf{t}_k = \left [ t_0, t_1, \dots, t_{p_\text{d} + N_\text{c}} \right]^{\text{T}}$ with constant interval $\Delta t$. The velocity $\mathbf{V}_i$, acceleration $\mathbf{A}_i$ and jerk $\mathbf{J}_i$ can be obtained by
\begin{equation}\label{equ:smoothnessofQ}
\mathbf{V}_i = \frac{\mathbf{Q}_{i+1} - \mathbf{Q}_{i}}{\Delta t}, \;\mathbf{A}_i = \frac{\mathbf{V}_{i+1} - \mathbf{V}_{i}}{\Delta t},\;\mathbf{J}_i = \frac{\mathbf{A}_{i+1} - \mathbf{A}_{i}}{\Delta t},
\end{equation}

Thus, the cost function can be rewritten as
\begin{equation}\label{equ:smoothness}
C_{\text{s}}+C_{\text{k}}+C_{\text{e}} = \omega_{\text{acc}}\sum_{i = 0}^{N_\text{c} - 2} \left \| \mathbf{A}_i \right \| ^2_2 + \omega_{\text{jerk}}\sum_{i = 0}^{N_\text{c} - 3} \left \| \mathbf{J}_i \right \| ^2_2
\end{equation}

\begin{figure}[htb]
	\centering
	\includegraphics[width=0.83\columnwidth]{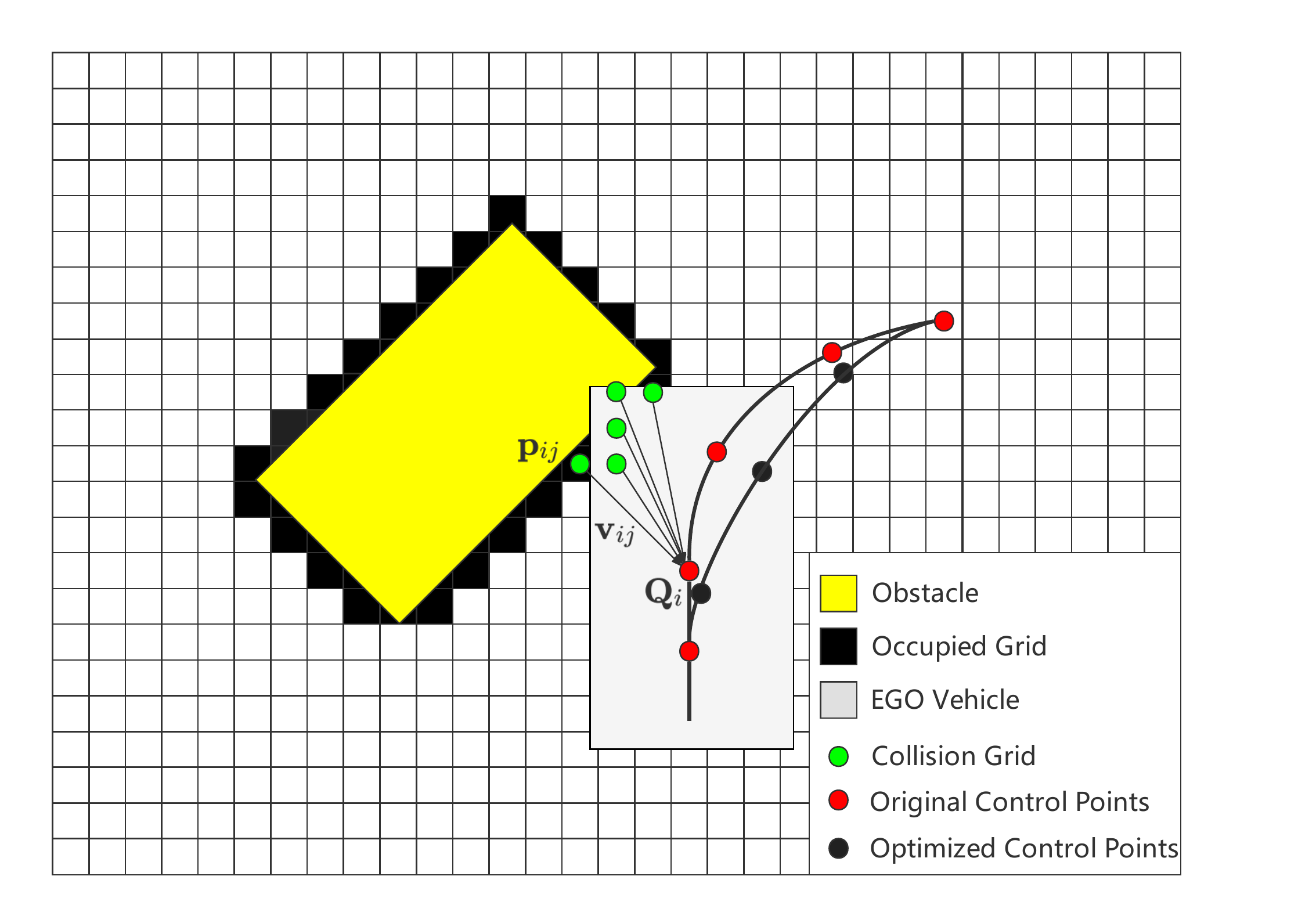}
	\caption{Generation of collision term in cost function}\label{fig::gradient}
\end{figure}

\subsection{Collision Avoidance}
The most difficult part of the low-level planner is the cost function with obstacle avoidance. In \cite{Zhou2020}, it is only applicable to objects with round or square outlines.
In a more general sense, we focus on the research of the planner itself. Therefore, we assume that the state of dynamic obstacles can be perfectly predicted. Thus, sampling the position of dynamic obstacles at a single time point can be treated as a static obstacle \cite{Zhao2020}. 
In a typical going straightforward scenario, as shown in Fig. \ref{fig::gradient}. we sample the position $\mathbf{P}_b(t_i)$ of B-spline trajectory on the time knot, $t_i \in \left \{t_{p_d}, t_1 \dots t_{p_d + N_\text{c} + 1}\right \}$, we check whether there are potential collisions. With the predictor we can get the position and state of the obstacles and the grid occupied by the obstacles can be obtained. Record the $j$-th potential collision grids nearby$\mathbf{P}_b(t_i)$ as $\mathbf{p}_{ij}$, then the gradient generation vector can be calculated by
\begin{equation}\label{equ:vector}
\mathbf{v}_{ij} = \mathbf{P}_b(t_i) -\mathbf{p}_{ij}.
\end{equation}

Theoretically, the closer the collision point, the greater the cost. A reasonable piecewise polynomial loss function is necessary, which is twice continuously differentiable. First, if the collision points are within a circle with the vehicle center as the center of circle and the diagonal of vehicle as the diameter, it will apply a higher penalty to the control points. Considering the error of control model and the grid size, we set an additional redundancy to keep collision-free, which apply a lower penalty. Finally, the piecewise polynomial function of collision avoidance can be represented by

\begin{equation}\label{equ:collision}
C_\text{c}(i,j) =\left \{
    \begin{array}{ll}
    0 & (d_{ij}<0)\\
    d_{ij}^3 & (0\leq d_{ij} <d_\text{r})\\
    3d_\text{f}d_{ij}^2 - 3d_\text{f}^2d_{ij}+d_\text{f}^3 &(d_\text{r} \leq d_{ij})
    \end{array}
\right.
\end{equation}
where $d_{ij} = d_\text{r} + d_\text{f} - \left \| \mathbf{v}_{ij}\right \|_2$, $d_\text{r}$ and $d_\text{f}$ are two important parameters of piecewise polynomials. $C_\text{c}(i,j)$ is the collision cost value of $\mathbf{P}_b(t_i)$ caused by collision gird $\mathbf{p}_{ij}$. The total collision cost of $\mathbf{P}_b(t_i)$ can be evaluated independently, {\textit{i.e.}}, $C_\text{c}(\mathbf{P}_b(t_i)) =  {\textstyle \sum_{i = 0}^{N_i - 1}C_\text{c}(i,j)}$, $N_i$ is the number of collision grid of $\mathbf{P}_b(t_i)$. Then the collision cost of entire trajectory can be evaluated by all control points.
\begin{equation}\label{equ:costOfQ}
{C}_\text{c} = \omega_{\text{c}}\sum_{i = 0}^{p_d + N_\text{c} } {C}_\text{c}(\mathbf{P}_b(t_i))
\end{equation}

For the gradient of collision term, it can be evaluated by the derivative of $C_\text{c}$ with respect of $\mathbf{Q}_i$ directly.

\begin{equation}\label{equ:gradient}
    \begin{aligned}
        \frac{\partial {C}_\text{c}}{\partial\mathbf{Q}_i} &= \sum_{i = 0}^{p_d + N_\text{c}}\sum_{j = 0}^{N_c - 1} \omega_{i,j} \frac{\mathbf{v}_{ij}}{\left \|\mathbf{v}_{ij}  \right \|_2 }
        \\
        & \cdot \left \{
        \begin{array}{ll}
        0 & (d_{ij}<0)\\
        -3d_{ij}^2 & (0\leq d_{ij} <d_\text{r})\\
        -6d_\text{f}d_{ij} + 3d_\text{f}^2 &(d_\text{r} \leq d_{ij})
        \end{array}
        \right.
    \end{aligned}
\end{equation}
where $\omega_{i,j}$ is the weight of $Q_j$ in time knot $t_i$.

\subsection{Trajectory Optimization}
The trajectory optimization problem considering collision avoidance can be modeled as an unconstrained optimization problem as (\ref{equ::lowlevel}), which uses some tricks to achieve approximate hard constraints.
\begin{equation}
\label{equ::lowlevel}
    \mathcal{P}_2: \;
	\underset{}{\mathop {\min_{\mathbf{Q}}. }} C_{\text{s}}+C_{\text{k}}+C_{\text{e}}+C_{\text{c}}
\end{equation}

We generate the control points of the B-Spline curve with the sampling points and states of the startpoint and endpoint of the reference trajectory. 
To ensure the continuity between the startpoint of the curve and the previous frame trajectory result, we cannot optimize the first three control points. However, due to the iteration of the planner, the last three control points can be optimized. Therefore, the number of control points must be greater than four. Since the objective function is higher than quadratic and quadratic terms dominate, the Hessian matrix can be used to accelerate the optimization process. In \cite{Zhou2020}, L-BFGS solver \cite{Boggs} has a good performance in unconstrained optimization, thus, we optimize the control points with L-BFGS solver.

When obstacles appear in the intersection, bypassing these obstacles may cause the vehicle to deviate from the feasible tunnel. This means that other vehicles need to consider this vehicle as an unexpected obstacle. Large obstacles may block the entire intersection, but this is not within the scope of our research.
After optimization, collision may still exist or the trajectory is not smooth enough, low-level planner can optimize again by increasing the weight of the corresponding term to make the trajectory meet the constraints, but in most cases, once is enough.

\section{Simulation Results}
\label{sec::simutaion}
In this section, we performed simulations in a typical four-lane road intersection scenario as shown in Fig. \ref{fig::intersection}, to demonstrate the performance of our proposed method. We compare it with the existing solutions in Section \ref{sec:Benchmark}.
We consider only the space-time resources within the intersection Conflict Area. The reference trajectory of the vehicle is given by Section \ref{sec::RTG} and \ref{sec::TBA}.

\begin{table}[htb]%
    \small
    \extrarowheight=2.5pt
    \arrayrulewidth=0.5pt
	\centering
	\caption{Calibration of Main Parameters}\label{tab::parameters}
	\begin{tabular}{| c | c | c | c |}
		\hline
		Parameters & Value & Parameters & Value \\
		\hline
		$W_\text{L}, L_\text{B}$ & $4\text{m}, 8\text{m}$
        & $d_{\text{b}}$ & $1.0\text{m}$ \\

        $\omega_2^{\text{ref}}, \omega_3^{\text{ref}},\omega_4^{\text{ref}},$ & $1,1,1$
        & $v_{\text{ref}}$ & $8\text{m/s}$ \\

        $dx$,$dy$ & $0.2\text{m}$
		& $dt$ & $0.2\text{s}$ \\

        $L_{\text{car}},\; W_{\text{car}}$& $4\text{m},2\text{m}$
	 	& $a_\text{r}^{\text{av}}$  & $0.8$  \\

        $R_{\text{long}},R_{\text{lat}}$ & $0.5\text{m},0.5\text{m}$
        & $\omega_{\text{acc}}, \omega_{\text{jerk}}$ & $5,1$ \\

        $\omega_{\text{d}}, \omega_{\text{w}}, \omega_{\text{sta}}$ & $1,5*10^{-1},1.0 $
        &  $\omega_{\text{c}}$ & $0.1$ \\

        $d_\text{r} ,d_\text{f}$ & $0.5\text{m},2.5\text{m}$
        & & \\

        \hline
	\end{tabular}%
\end{table}

\subsection{high-level planner}

\begin{figure}[htb]
	\centering
	\includegraphics[width=0.83\columnwidth]{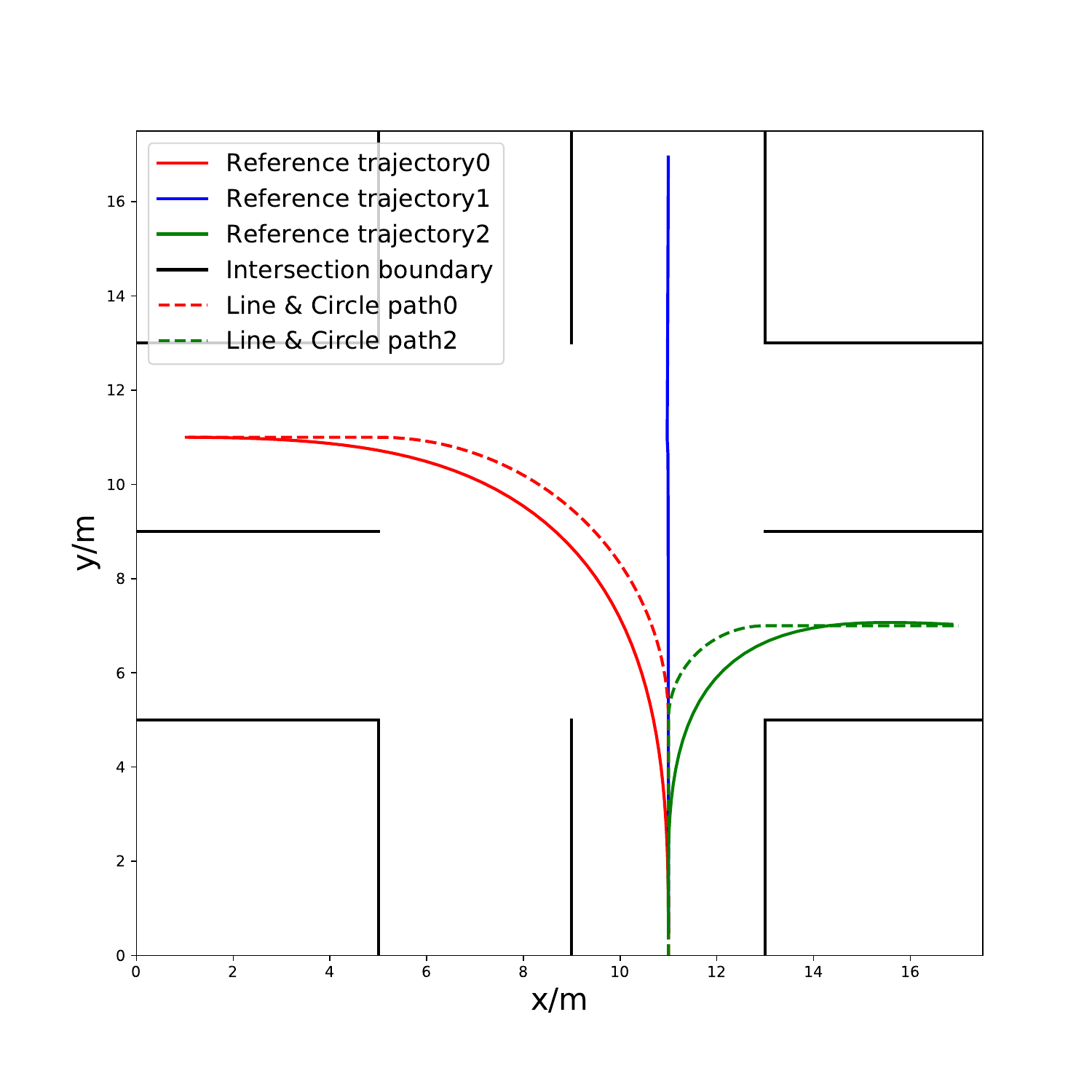}
    \vspace{-2em}
	\caption{The reference path and line \& circle path}
    \label{fig::referencepath}
\end{figure}

The simulation parameters are listed in Table. \ref{tab::parameters}. Fig. \ref{fig::referencepath} shows the reference trajectory provided by the Reference Trajectory Generator, which avoids the flaw of line \& circle curve paths and improves smoothness and feasibility. Theoretically, the Reference Trajectory Generator is scalable in multi-lane road intersections.

\begin{figure}[htb]
	\centering
	\includegraphics[width=0.83\columnwidth]{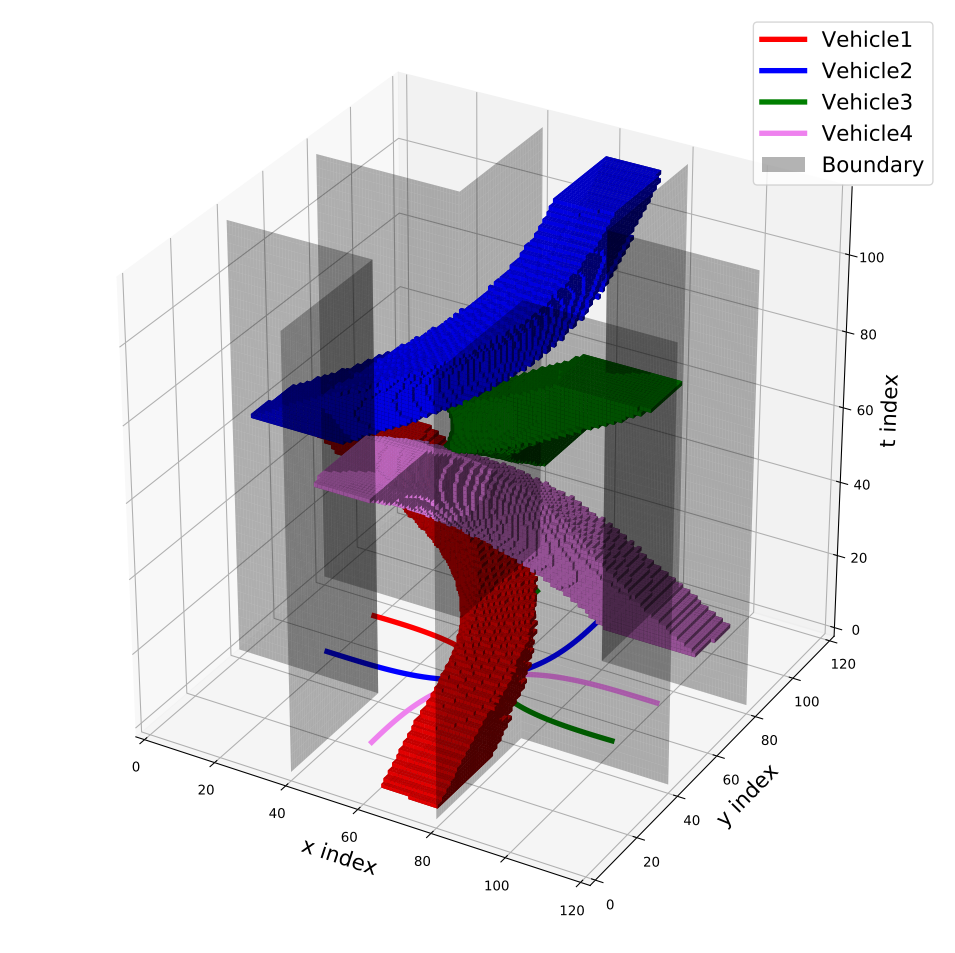}
    \caption{The reference trajectories and resource allocation for four left-turning CAVs from different lanes}
    \label{fig::resourceOf4}
\end{figure}

\subsubsection{Space-Time Resource Allocation}
For four left-turning CAVs from different roads, theoretically, it is a reasonable solution that CAVs enter the intersection in sequence. The same strategy is given in our experiments.
Fig. \ref{fig::resourceOf4} shows the reference trajectories and resource allocations of the CAVs in this scenario, we only set $dt = 0.05\text{s}$ in particular simulation for better illustration of space-time resource blocks.

In this result, the resource blocks of different CAVs touch on the t-axis, which indicating that the CAVs pass through the road intersection without collision. This essentially shows that single vehicle finds the optimal solution, {\textit{i.e.}} the fastest solution to get through the road intersection, subject to the order of passage provided by the high-level planner.

\begin{figure}[htbp]
\centering
\subfloat[Comparison of our proposed method and CS strategy in ST coordinate system]{
\includegraphics[width=0.83\linewidth]{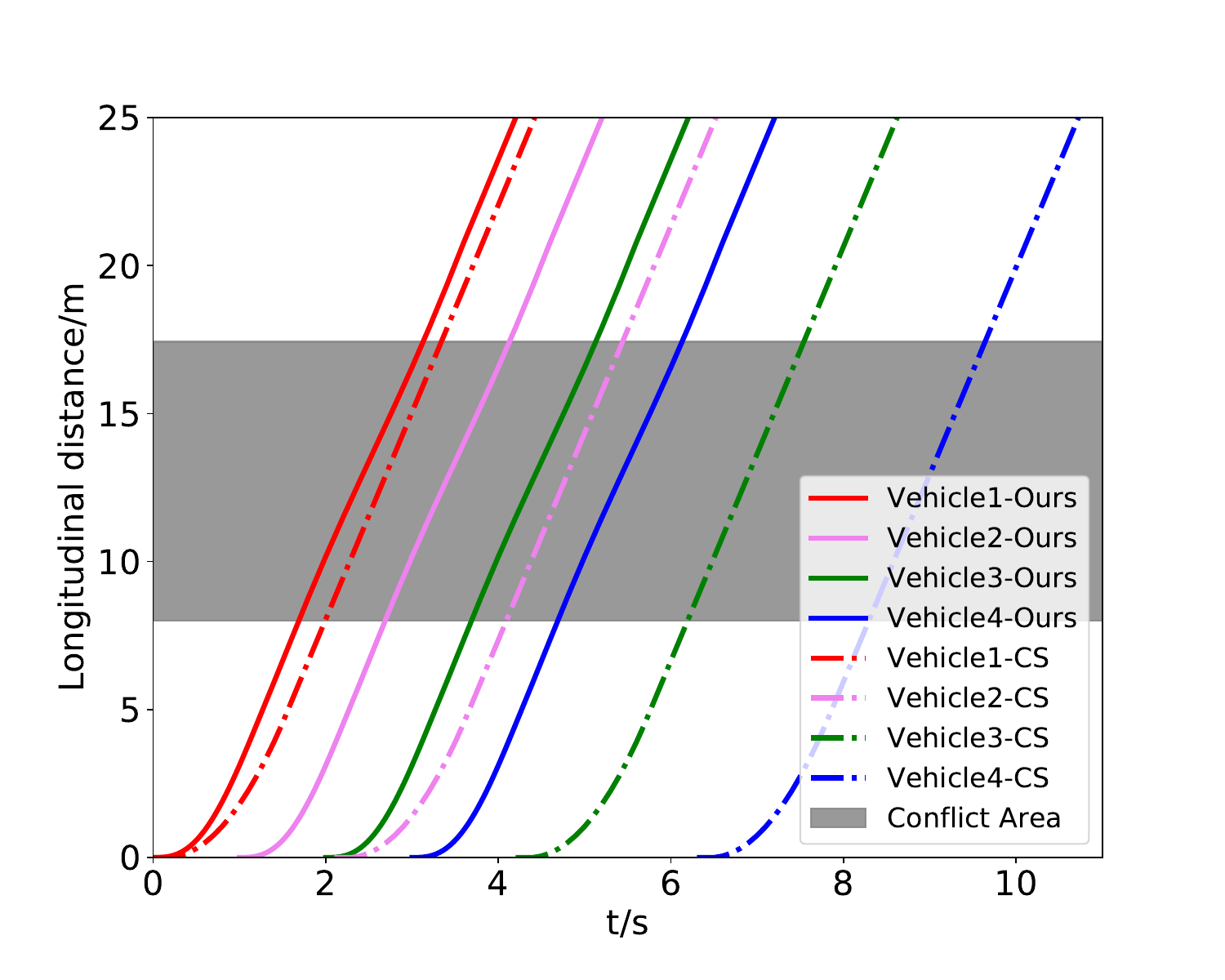}
\label{fig::OUR_CS}
}
\\
\subfloat[Comparison of our proposed method and STRS strategy in ST coordinate system]{
\includegraphics[width=0.83\linewidth]{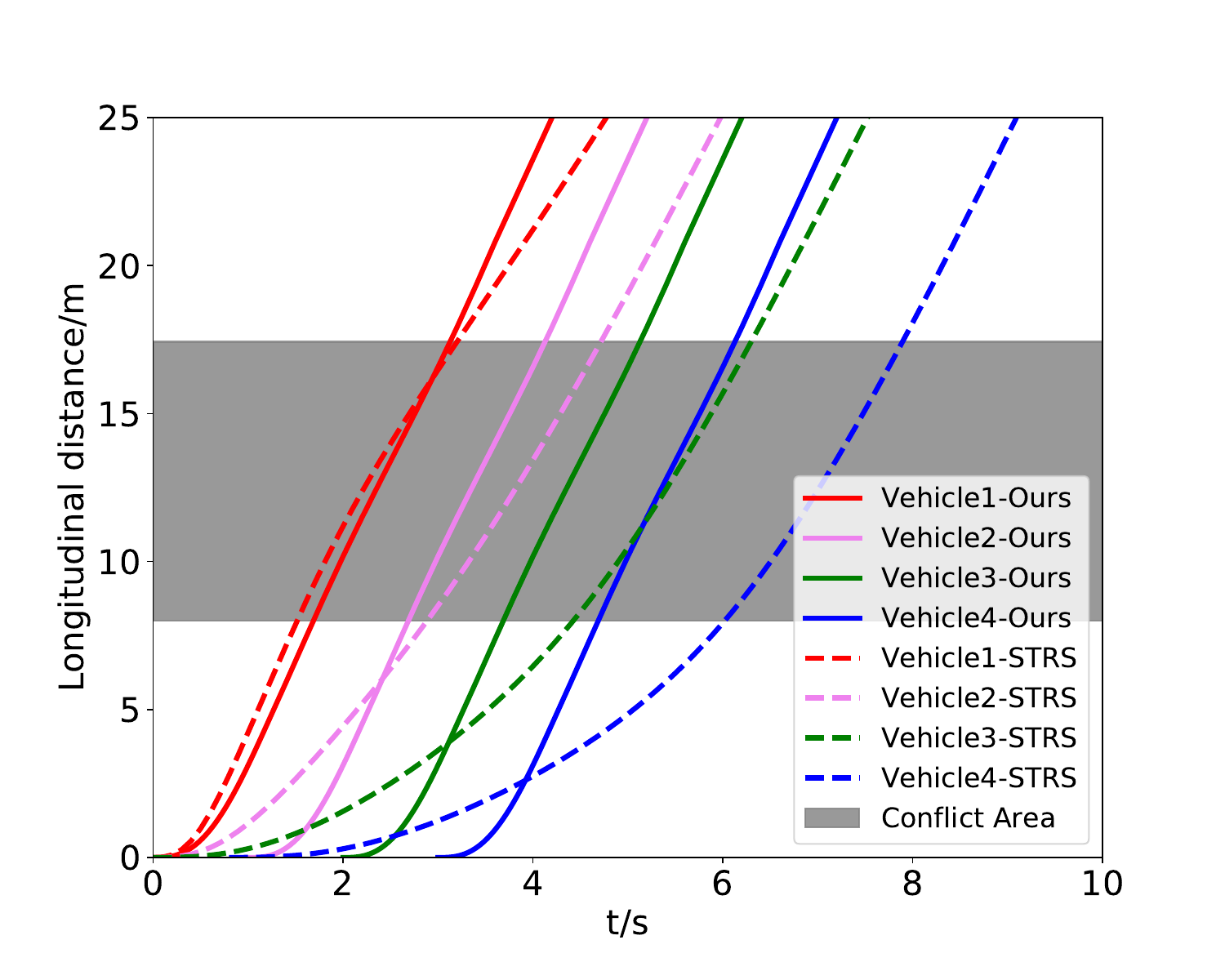}
\label{fig::OUR_STRS}
}
\caption{Comparison of different methods in ST coordinate system}
\label{fig::STcomparision}
\end{figure}

\subsubsection{Speed Profile Analysis}\label{sec::spa}
In a typical four left-turn CAVs scenario, Fig. \ref{fig::STcomparision} represents the output of our proposed method compared with CS and STRS in ST coordinate system of Fren\'et Frame \cite{Werling2010}, respectively. In Fig. \ref{fig::STcomparision}\subref{fig::OUR_CS}, since the solution of CS strategy is that the road intersection can be occupied by only one vehicle at the same time, the latter vehicle must enter the intersection after the former one has left, this strategy leads to a less efficient intersection passage.

In Fig. \ref{fig::STcomparision}\subref{fig::OUR_STRS}, unlike the CS and our proposed approach, the STRS strategy adjusts the speed profile by DP+QP to find the optimal solution. Since the goal of the STRS strategy is to find the optimal solution of comfort and fuel efficiency without collision for each vehicle, which is often not optimal for pass throughput.

The key issue of our proposed strategy is to ensure that the space-time resource blocks do not overlap, to guarantee safety. While the trajectory comfort could be guaranteed by the reference trajectory. Therefore, combining the results of the three strategies above, our proposed method improves the traffic throughput compared to the CS strategy, but the comfort level is not as high as STRS.

\begin{figure}[htb]
	\centering
    \subfloat[Resource allocation without waiting time term]{
	\includegraphics[width=0.75\columnwidth]{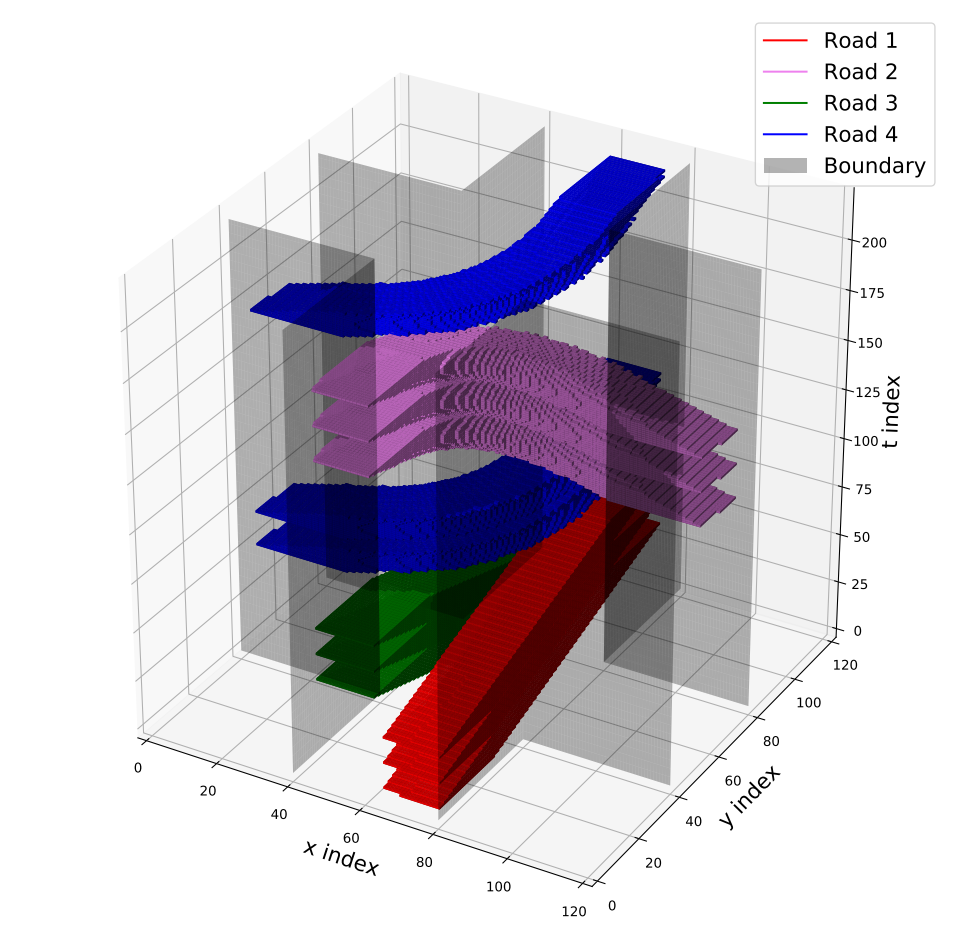}
    \label{fig::resourceOf12_1}
    \vspace{-1em}
    }
    \\
    \subfloat[Resource allocation with waiting time term]{
	\includegraphics[width=0.75\columnwidth]{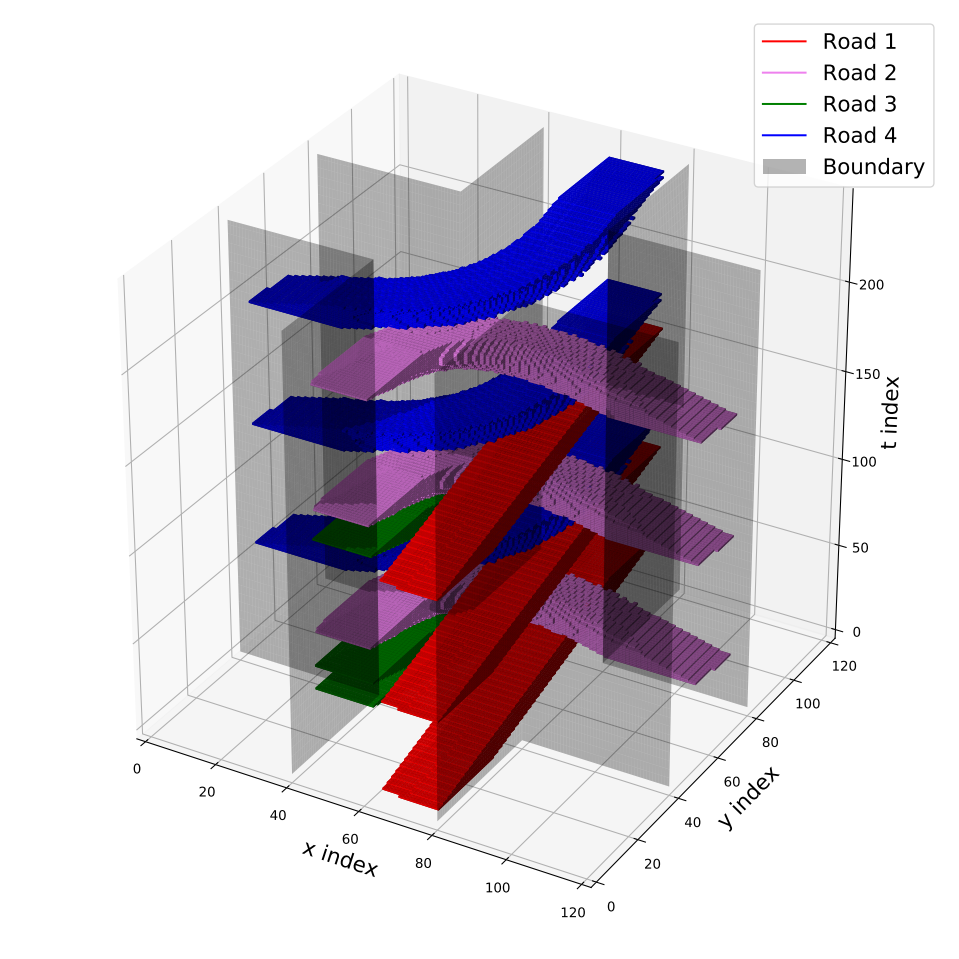}
    \label{fig::resourceOf12_2}
    \vspace{-1em}
    }
    \caption{Illustration of resource allocation in the extreme scenarios}
    \label{fig::resourceOf12}
\end{figure}

\subsubsection{Waiting Time Term}
We set up an extreme simulation scenario where there are three CAVs going straight on Road 1 and Road 3 respectively, and three CAVs turning left on Road 2 and Road 4 respectively. Fig. \ref{fig::resourceOf12}\subref{fig::resourceOf12_1} shows the resource allocation results without considering the waiting time term, where the resources allocation for CAVs from different roads are distinguished by different colors.
Because CAVs going straight have shorter distances and can move out the intersection earlier, the high-level planner prefer to raise the priority of CAVs going straight, which would lead to unfairness.
In Fig. \ref{fig::resourceOf12}\subref{fig::resourceOf12_2}, with the introduction of the waiting time term to priority, the scheduler will avoid the situation that long waiting time of the first vehicle in road queue and ensure the fairness of CAVs with different maneuvers in different roads.

\subsubsection{Total Passing Time}

\begin{figure}[htb]
	\centering
	\includegraphics[width=0.83\columnwidth]{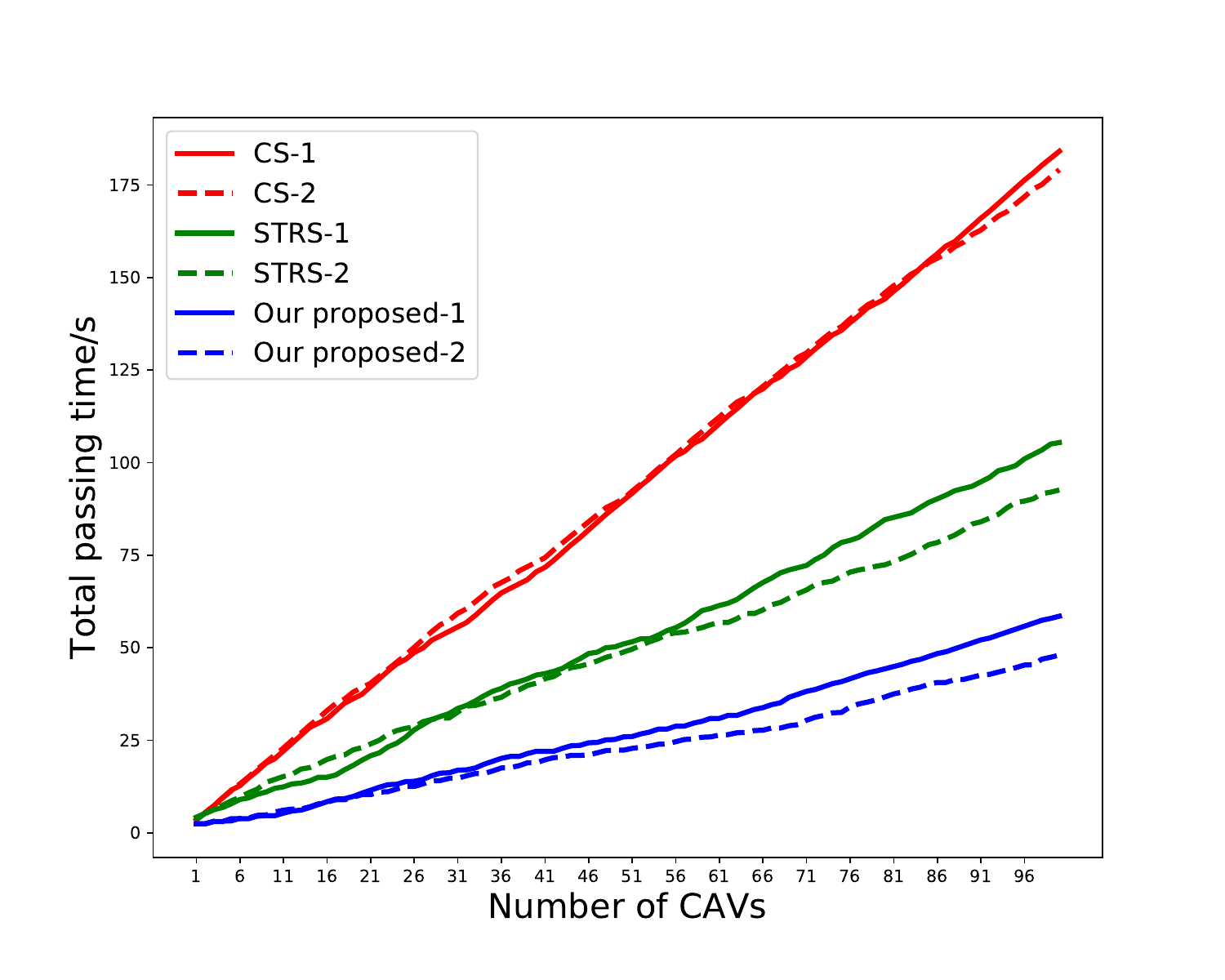}
    \vspace{-1em}
    \caption{The total passing time of the three strategies}
    \label{fig::passingTime}
    \vspace{-1em}
\end{figure}

Due to the different objectives of the High-Level coordination method, it mainly faces the contradiction between comfort and traffic efficiency. Therefore, the actual performance of the algorithm cannot be explained by directly comparing the total passing time of same traffic flow. We generate two kinds of traffic flows to compare the ability of different methods to tap the potential of the road intersection in the face of different traffic flows. Traffic flow 1 consists of 50\% left-turning CAVs, right-turning CAVs and straight-going CAVs accounted for 25\% each. Traffic flow 2 consists of 50\% straight-going CAVs, 25\% right-turning CAVs and 25\% left-turning CAVs. Fig. \ref{fig::passingTime} shows the total passing time of CS, STRS, and our proposed approach versus the number of vehicles when facing different traffic flows.

Just like the conclusion of \ref{sec::spa}, due to the waste of resources, CS strategy can not make the most of the resources at the crossroads and has the lowest traffic efficiency of the road intersection. Because in the face of different traffic flows, the two total passing time did not widen the gap.
The STRS strategy try to find the optimal solution of comfort and fuel efficiency, which means that the total passing time of STRS strategy is not always the optimal. However, this method can greatly realize the potential of intersections in the face of traffic flow with more straight traffic.
Similar to STRS, our proposed approach can also ensure the efficiency of intersections in the face of different traffic flows. which sacrifices some comfort and achieves an increase in intersection throughput.
Nonetheless, a possible corollary is that modifications to the optimization objective in STRS strategy can theoretically lead to an optimal solution for the passage efficiency, since the STRS strategy searches all speed profiles at a certain resolution iteratively.

\begin{figure}[htb]
	\centering
	\includegraphics[width=0.83\columnwidth]{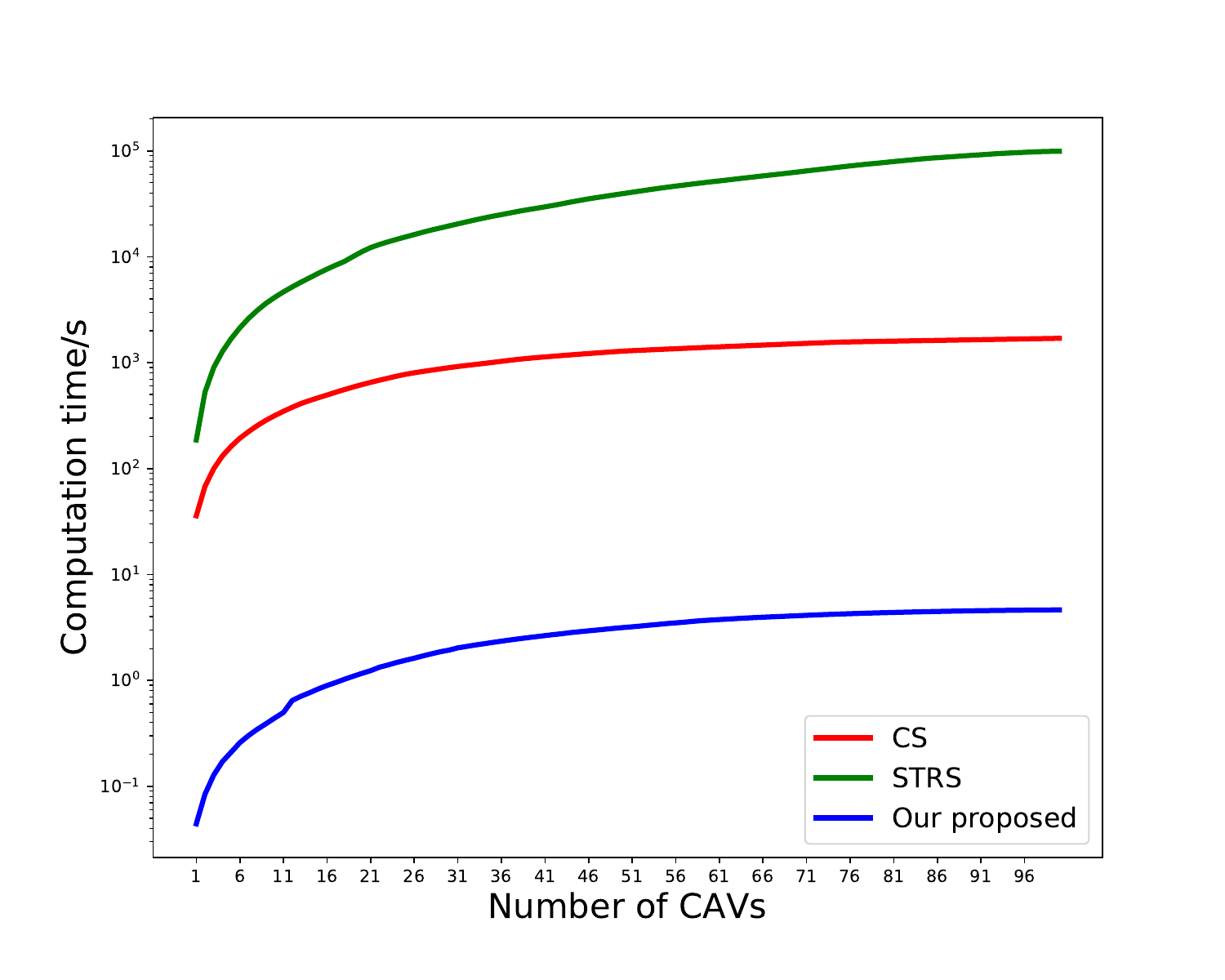}
    \vspace{-1em}
    \caption{The computation time of the three strategies}
    \label{fig::timeCost}
\end{figure}

\subsubsection{Computation Efficiency}
We conducted comparative experiments using the source code provided by \cite{zhang2021}. In the same traffic flow scenario, Fig. \ref{fig::timeCost} demonstrates a phenomenon: the computation time performance of our proposed strategy outperforms other strategies for reference only. This can be attributed to the fact that the CS strategy uses a Quadratic Programming + Linear Programming solving approach, which results in a much lower computational complexity compared to the DP+QP strategy used in STRS. However, it's worth noting that the CS source code employs velocity sampling points as the optimization target, leading to large-scale QP problems and an unacceptable computational efficiency.

We believe that with careful optimization of the source code, the CS strategy might achieve computational efficiency similar to that of our proposed method, as CS does not need to consider a large number of spatiotemporal resource blocks but requires considering a considerable number of velocity curve sampling points, owing to the nature of the CS strategy. However, even with optimization, the STRS strategy struggles to achieve satisfactory computational efficiency due to its inherent high algorithm complexity.

Therefore, it is important to highlight that the comparative experiments regarding computational efficiency are intended for reference purposes only. They do not imply that the three algorithms exhibit such a significant disparity in computational consumption during actual deployment, as the quality of code optimization may vary significantly between different implementations.

\subsubsection{The relationship between redundancy and traffic efficiency}

\begin{figure}[htb]
    \centering
    \includegraphics[width=0.83\columnwidth]{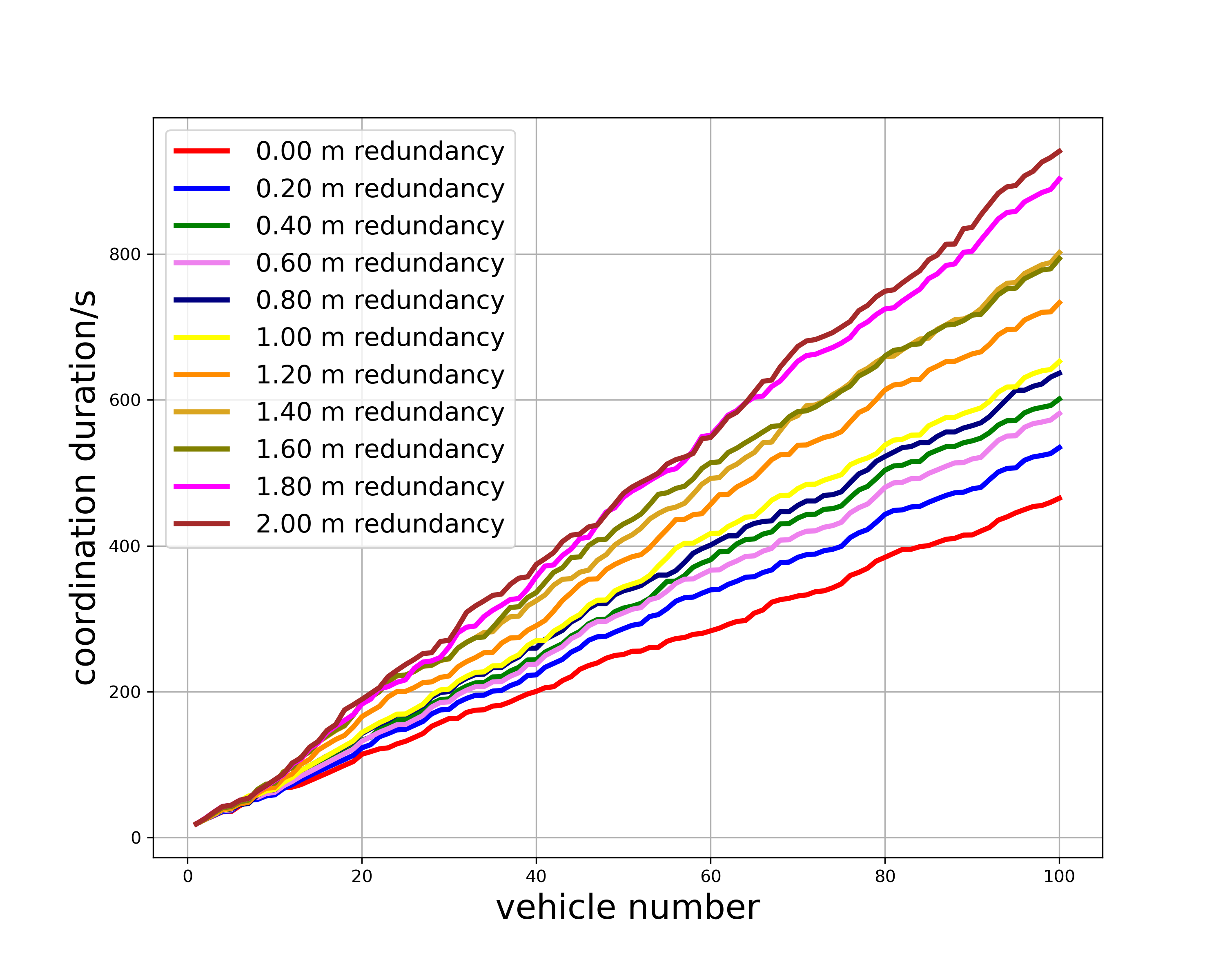}
    \caption{Coordination duration for each vehicle with different redundancy}\label{fig::time_cost_veh_num}
\end{figure}

\begin{figure}[htb]
    \centering
    \includegraphics[width=0.83\columnwidth]{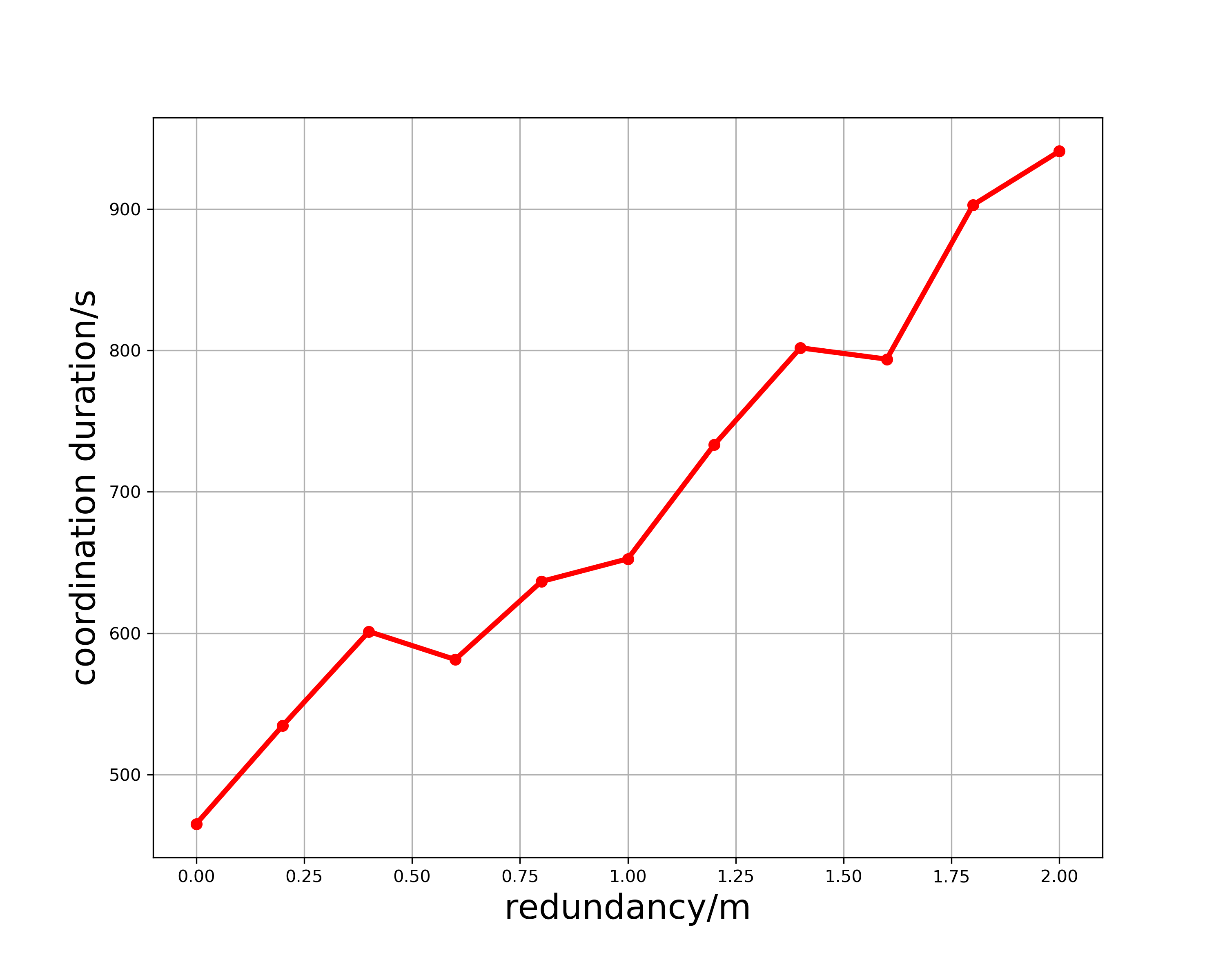}
    \caption{Coordination duration for 100 vehicles with different redundancy}\label{fig::time_cost_redundancy}
\end{figure}

We performed experiments using the same high-level planner parameters as outlined in Table. I. Redundancy size was set from $0$m to $2$m with an interval of $0.2$m. The experiments were conducted with 100 vehicles in a random traffic flow scenario, where vehicles were uniformly distributed, randomly entering and departing from different lanes (excluding the entering lane). We recorded the entry time of each vehicle at the intersection.
Fig. \ref{fig::time_cost_veh_num} illustrates the entry time of the 100 vehicles at the intersection for different redundancy sizes, while Fig. \ref{fig::time_cost_redundancy} demonstrates the variation in scheduling efficiency for different redundancy sizes. By comparing the data shown in the figure, we observed that the overall scheduling efficiency decreases significantly as the redundancy size increases. However, different redundancy sizes can cause changes in the baseline of vehicle priority, so without adjusting the corresponding parameters, there might be a slightly higher throughput efficiency with a slightly larger redundancy. The essential reason is that we prioritize optimal scheduling priority in each round, rather than optimizing traffic efficiency. Additionally, it is not realistic to consider scheduling efficiency alone as the sole optimization objective also the reason for the change. Based on previous research experience \cite{zhang2021, Yao2023, Wu2022}, it is necessary to transform it into a sequence optimization problem, which also implies the utilization of receding horizon control. Nevertheless, larger redundancy provides the scheduling system with higher robustness to deal with unexpected events such as control system errors, positioning system errors, and unexpected obstacles. Therefore, we confidently conclude that the redundancy size represents the trade-off between scheduling system efficiency and robustness.

\begin{table}[htb]%
    \small
    \extrarowheight=2.5pt
    \arrayrulewidth=0.5pt
	\centering
	\caption{Calibration of Main Parameters}\label{tab::parameters2}
	\begin{tabular}{| c | c | c | c |}
		\hline
		Parameters & Value & Parameters & Value \\
		\hline
        $\Delta t$& $1.0\text{s}$
	 	&  $d_{\text{r}}$ & $2.5\text{m}$\\
        $L_{\text{car}}, W_{\text{car}}$ & $4.0\text{m}, 2.0\text{m}$
		& $d_{\text{f}}$ & $0.6\text{m}$ \\
        $\omega^{\text{acc}},\omega^{\text{jerk}}, \omega^{\text{c}}$ & $1.0, 1.2, 1.0$
        & $R_{\text{long}},R_{\text{lat}}$ & $0.4\text{m},0.4\text{m}$ \\
        \hline
	\end{tabular}%
\end{table}

\begin{figure*}[htb]
	\centering
	\includegraphics[width=2.4\columnwidth]{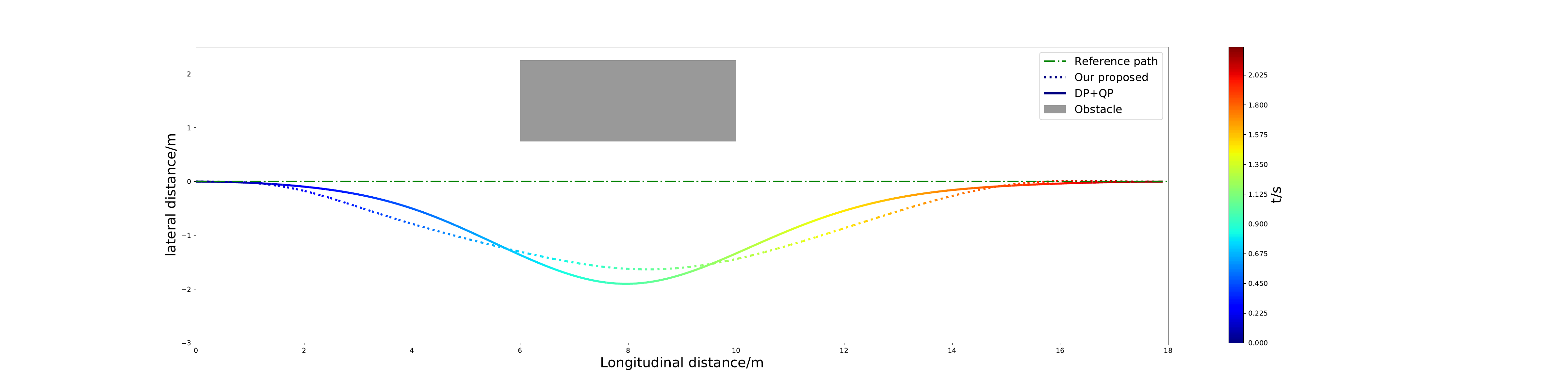}
    \caption{Comparison of simulation results between DP+QP \cite{zhangicas2021} and our proposed solution in Fren\'et Frame}
    \label{fig::localTraj}
\end{figure*}

\begin{figure}[htb]
	\centering
    \vspace{-1em}
	\includegraphics[width=0.83\columnwidth]{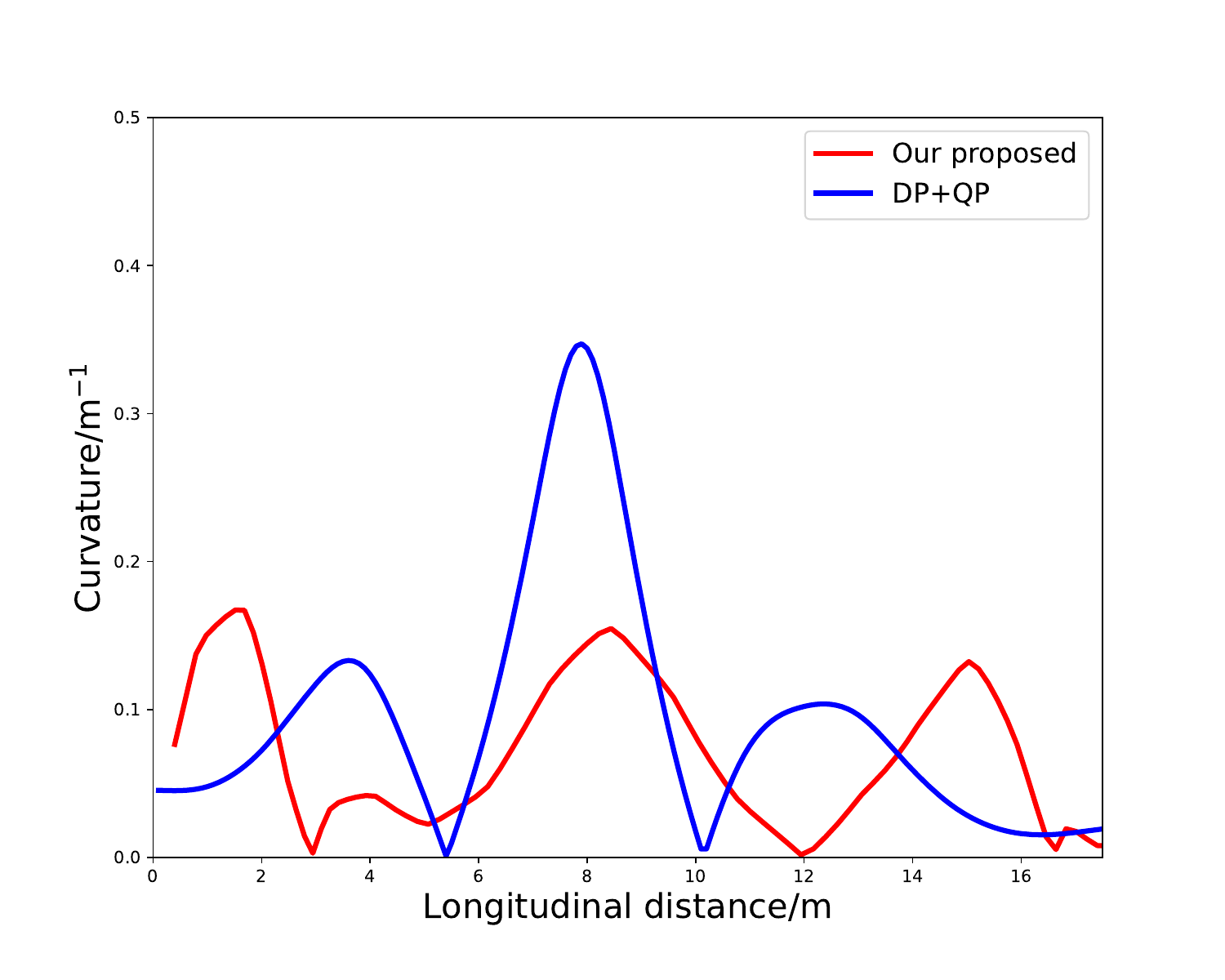}
    \caption{Comparison of curvature between DP+QP and proposed solution.}
    \label{fig::curvature}
    \vspace{-1em}
\end{figure}

\subsection{low-level planner}

\subsubsection{Trajectory Analysis}
Fig. \ref{fig::localTraj} shows a simulation results comparison between our proposed and the DP+QP \cite{zhangicas2021, zhang2021} strategy. The gray area represents an obstacle projected into the Fren\'et Frame, the solid line represents the result of DP+QP strategy, and the dashed line represents the simulation result of our proposed method. The simulation parameters are listed in Table. 2. We observed that both trajectories were effective in obstacle avoidance. However, the obstacle avoidance strategy of DP+QP, which is based on a circular bounding box model, requires a larger space, leading to increased safety redundancy and decreased efficiency in high-level planner. Additionally, when examining the curvature profile in Fig. \ref{fig::curvature}, the trajectories generated by the low-level planner of DP+QP lack smoothness. The curvature curve exhibits significant fluctuations, and high curvatures often approach the vehicle's kinematic limitation, resulting in lower comfort. On the other hand, our proposed algorithm generates trajectories with smaller and smoother curvature profiles, improving controllability and predictability in direction changes. This smoothness enhances the comfort of the vehicle, reduces wheel vibrations, and ensures adherence to the constraints of the vehicle's kinematic model \cite{Wang2022}.

\subsubsection{Dynamic obstacle avoidance} 
\begin{figure}[htb]
    \centering
    \includegraphics[width=1.2\columnwidth]{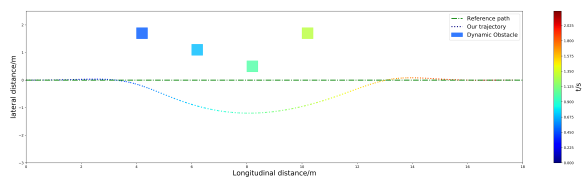}
    \caption{Trajectory planner in dynamic obstacle scenario}\label{fig::dynamic_traj}
\end{figure}

Essentially, our planner can extend obstacle avoidance from static obstacles to dynamic case. To demonstrate this, an additional experiment is finished to prove the ability of our proposed planner to evade dynamic obstacles, as shown in Fig. \ref{fig::dynamic_traj}. In the experiment, the predicted trajectories of obstacles were predefined, and we sampled the predicted trajectories of the obstacles at each time knot. For clarity, we only show the relevant nodes affecting the trajectory. From the figure, we can observe that our planner can effectively dodge dynamic obstacles, just as it would with static obstacles.

\subsubsection{Computation Time}
Computational time consumption is also an important consideration for the local planner, as faster computation means vehicles have more time to respond to unexpected events. Therefore, we compared the computation time between the DP+QP strategy and our proposed method to illustrate this. Typically, in practical autonomous vehicles, the DP+QP strategy can be accelerated using multithreading. Although our proposed low-level planner could also benefit from multithreading, we deliberately used single-thread computation for consistency. In the context of intersection scenarios, excessive deceleration of vehicles leads to intersection blockage. As a result, we compared our proposed method to the path planners in \cite{zhang2021} and \cite{fan2018}. In a typical obstacle avoidance scenario, as shown in Fig. \ref{fig::localTraj}, our method only requires $19.13ms$ to complete one frame of planning, including trajectory safety checks. We conducted experiments using default parameters provided in the referenced papers. Due to the exponential complexity, the path planners took several hundred milliseconds to compute, with a result of $803ms$ in our experiments. Therefore, our proposed method demonstrates excellent operational efficiency.

\section{Laboratory Experiments}
\label{sec::LabExp}

Finally, in order to verify the feasibility and robustness of our proposed double-level coordination strategy, we used four autonomous mobile robots with robot operating systems (ROS) \cite{2009ROS} to complete the laboratory experiments. We used the desktop computer to run the high-level planner, distributed the results of the high-level planner to the mobile robot through ROS network communication, and the low-level planner was completed with the mobile robot onboard computers. In order to conform to the reality, we mainly adjusted the experimental parameters: $dx = dy = 0.05\text{m}$, $\Delta t = 0.1\text{s}$, $L_{\text{car}} = 0.44\text{m}$, $W_{\text{car}} = 0.34\text{m}$.

\begin{figure}[htb]
	\centering
    \vspace{-1em}
	\includegraphics[width=0.83\columnwidth]{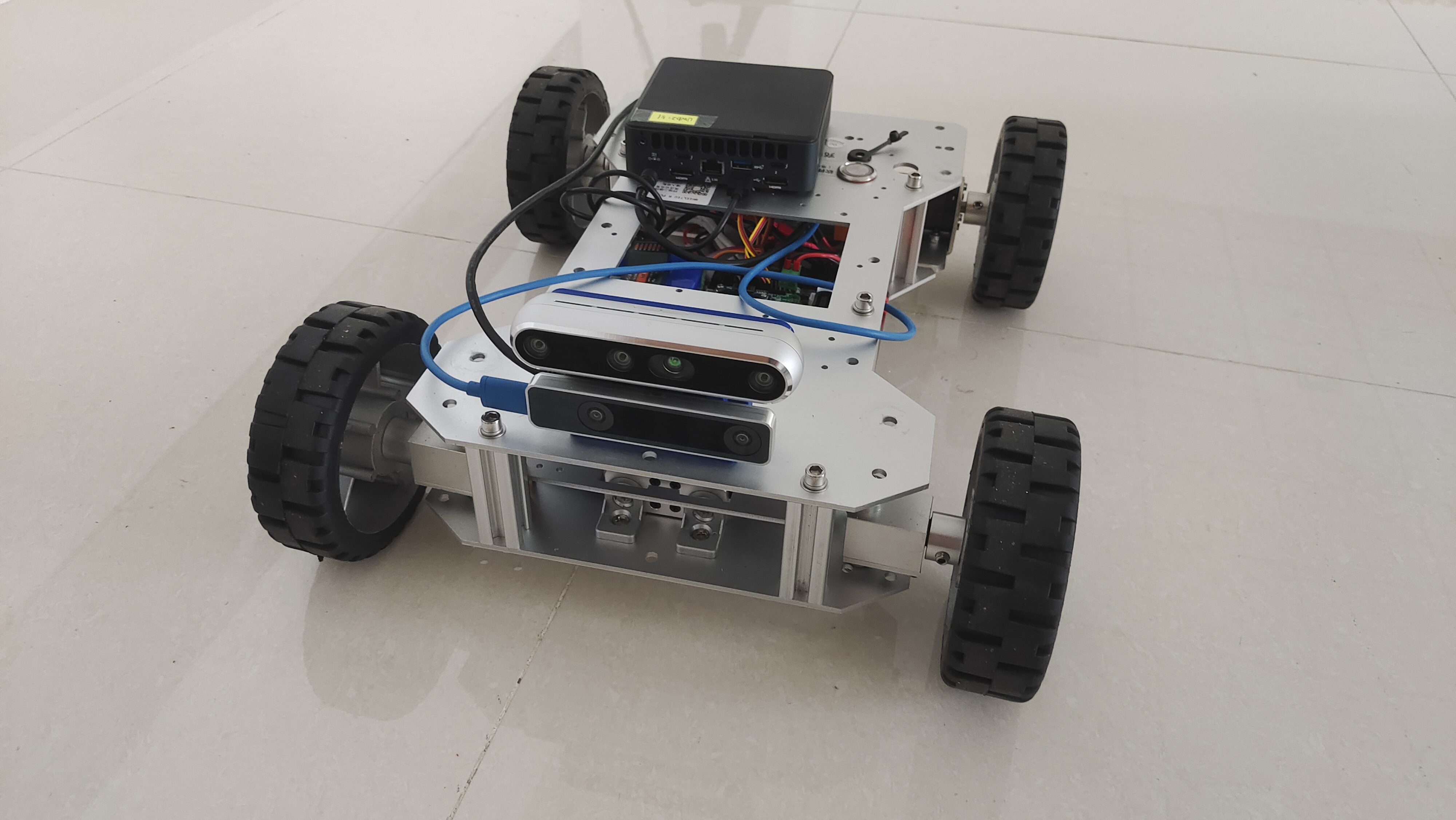}
    \caption{Autonomous mobile robots used in experiments}
    \label{fig::robot}
\end{figure}

Fig. \ref{fig::robot} shows the mobile robot we used, which is equipped with Intel T265 tracking camera and D455 depth camera.
In this experiment, we only used T265 for online localization, and perceptual results were generated offline. In addition, we use the Intel NUC microcomputer as the onboard computer, and the sufficient computing resources enable the platform to ensure real-time performance.

\subsection{high-level planner experiment \uppercase\expandafter{\romannumeral1}}

\begin{figure}[htb]
	\centering
	\includegraphics[width=1\columnwidth]{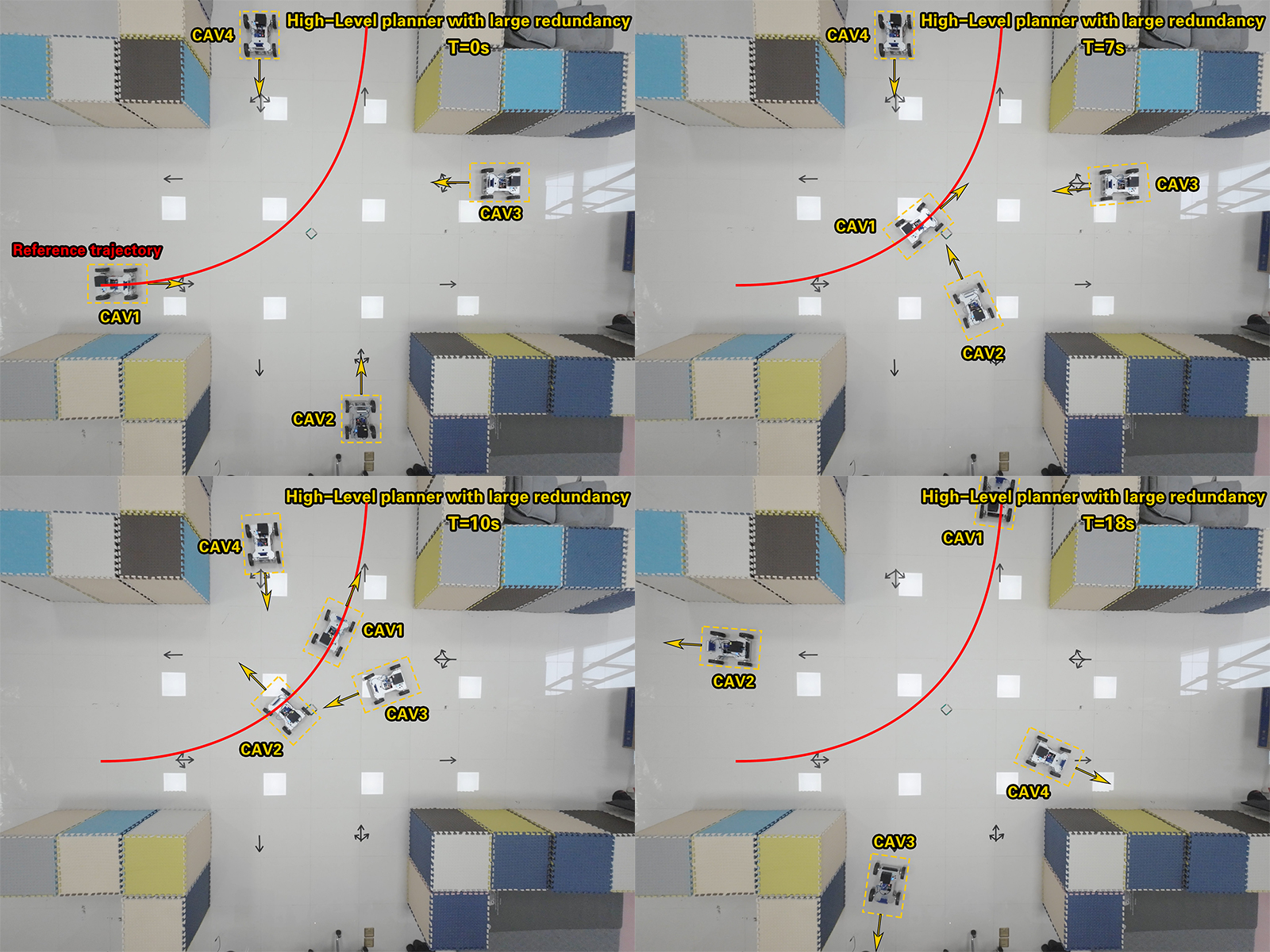}
    \caption{high-level planner laboratory experiments with large redundancy}
    \label{fig::test1}
\end{figure}

Fig. \ref{fig::test1} shows a typical laboratory road intersection scenario. In this experiment, four mobile robots all adopt a left turn strategy. We use large safety redundancy and fast reference speed, {\textit{i.e.}}, $R_{\text{long}} = R_{\text{lat}} = 0.4\text{m}$, $v_{\text{ref}} = 0.4\text{m/s}$. In this experiment, due to the large safety redundancy, CAVs in the diagonal roads cannot enter the intersection simultaneously, thus, the CAVs need to enter the intersection one by one. We plotted the reference path of CAV1 with red curve to show the error of the control system.
With proper security redundancy, our proposed can realize the throughput potential of the intersection without collision.

\subsection{high-level planner experiment \uppercase\expandafter{\romannumeral2}}

\begin{figure}[htb]
	\centering
	\includegraphics[width=1\columnwidth]{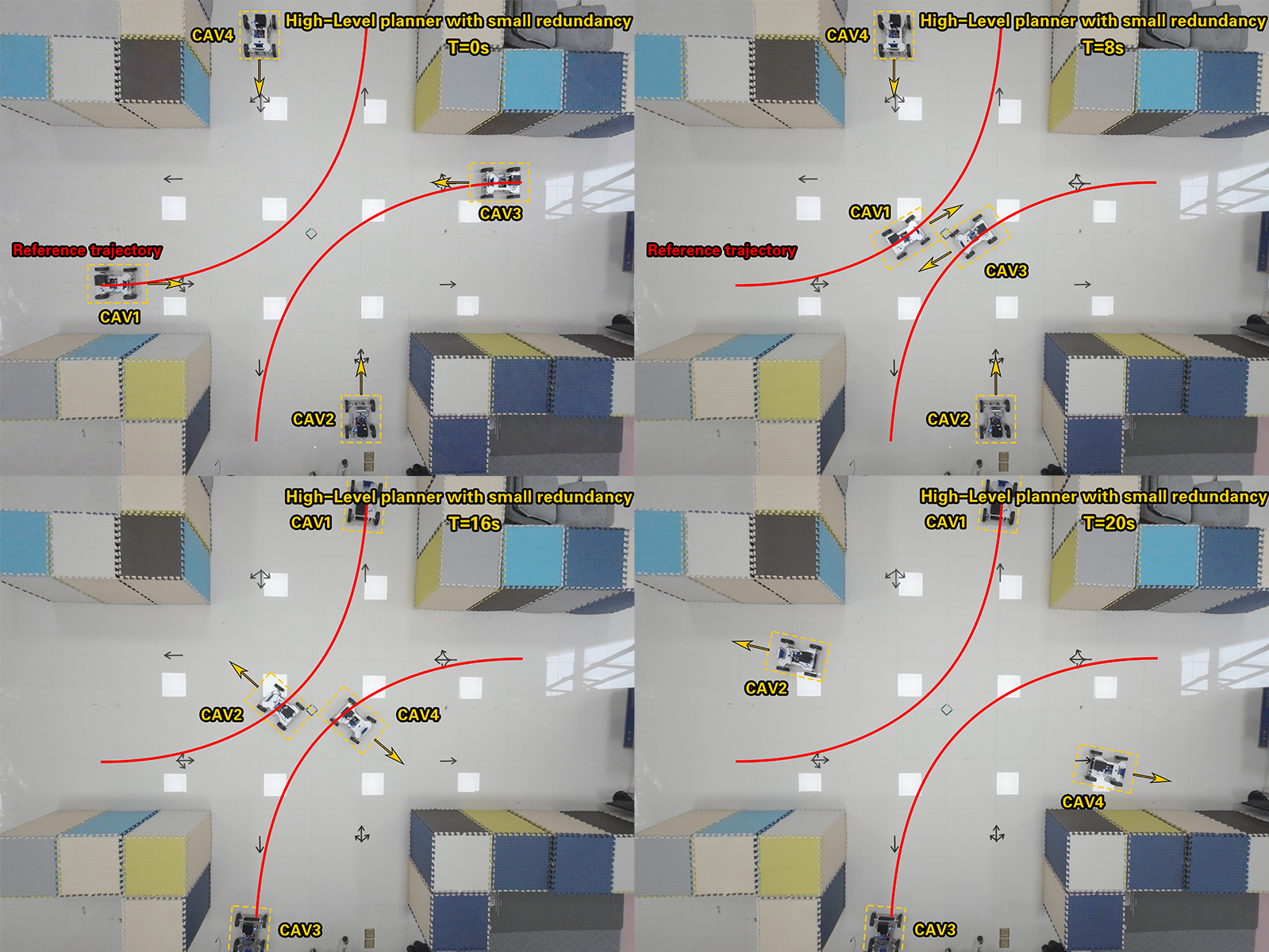}
    \caption{high-level planner laboratory experiments with small redundancy}
    \label{fig::test2}
\end{figure}

In Fig. \ref{fig::test2}, We reduce security redundancy, {\textit{i.e.}}, $R_{\text{long}} = R_{\text{lat}} = 0.1\text{m}$. We plotted the reference paths of CAV1 and CAV3 with red curves, their reference trajectories don't conflict, thus, the CAVs from the diagonal roads can enter the road intersection simultaneously.

\subsection{Double-level coordination strategy experiment}
\begin{figure}[htb]
	\centering
	\includegraphics[width=1\columnwidth]{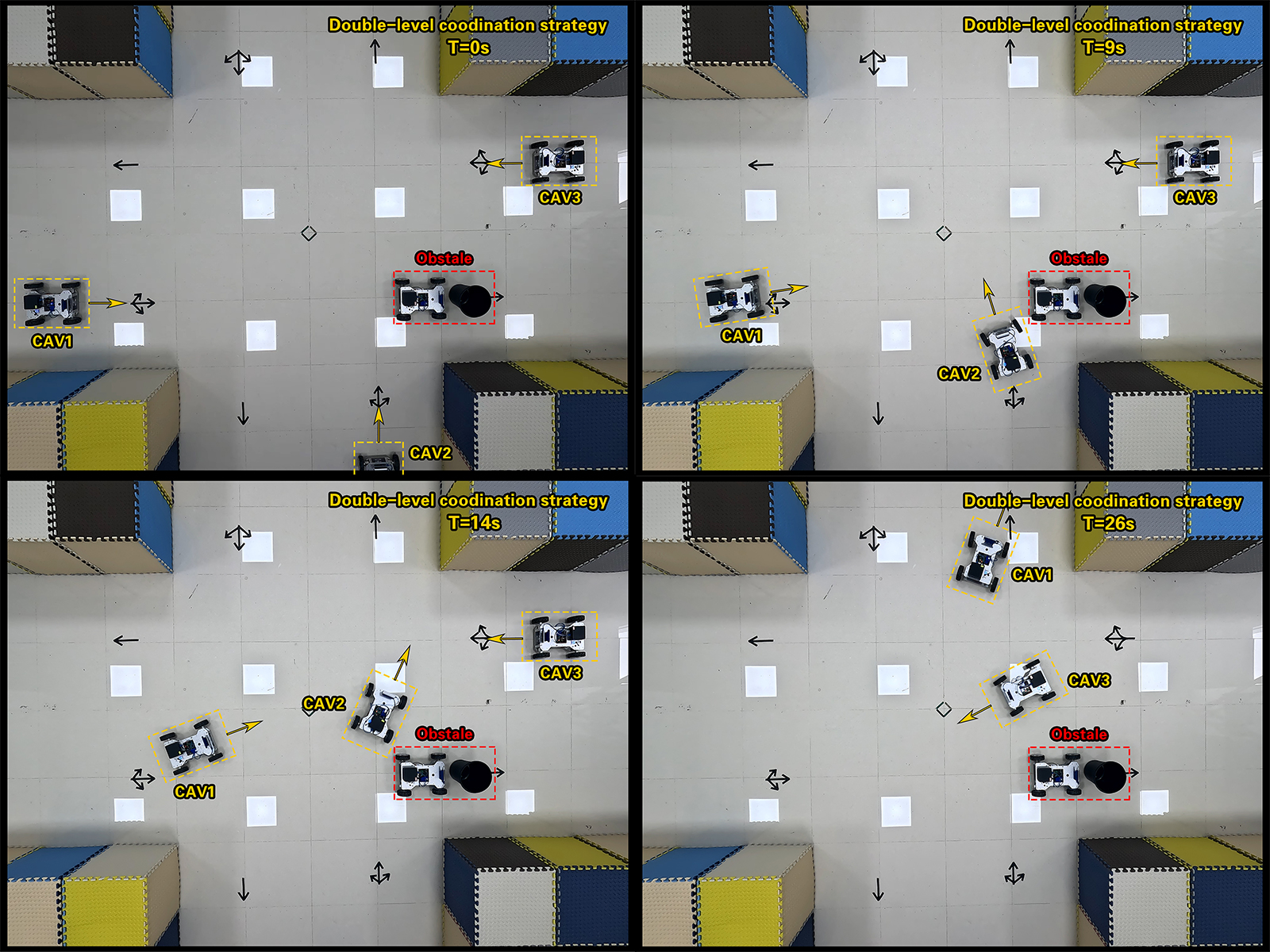}
    \caption{Double-level coordination strategy experiment with unexpected obstacle in intersection}
    \label{fig::test3}
\end{figure}

With unexpected obstacles in the intersection or a large control error, coordination strategy requires the intervention of the low-level planner. In this laboratory experiment, we adopt relatively lower reference speed ($v_{\text{ref}} = 0.2\text{m/s}$) and same redundancy with high-level planner experiment \uppercase\expandafter{\romannumeral1}. A static obstacle is arranged in the intersection, which is marked by red box. As a result of this obstacle, the reference trajectory of the CAV2 is exposed to potential collision. After spotting the obstacle, CAV2 generates the collision free trajectory base on the result of high-level planner by low-level planner. Fig. \ref{fig::test3} shows the results of the experiment.

\section{Conclusion}
\label{sec::Conclusion}
In this paper, We try to give a tightly coupled two level coordination framework for CAVs at non-signalized road intersections. Aiming at the computational complexity at the high-level planner, we decouple the original problem into sequential optimization ones, to generate the available space and time resource blocks, namely feasible tunnels, for each vehicle. Furthermore, a low-level planner is also given under the constraints from FTs, to avoid potential collisions with unexpected non-CAV obstacles. We provide numerical results via simulations and laboratory experiments, which could achieve coordination within millisecond level.
Although the uncertainties during coordination could be accommodated by the redundancy design of FTs, it is still prone to severe errors from sensing, control and communications. The robustness issue of this two-level coordinator will be discussed in our future work.

\bibliographystyle{IEEEtran}
\bibliography{myref}

\end{document}